%% file: main.tex
\documentclass[10pt,twocolumn,letterpaper]{article}

\usepackage{cvpr}      


\definecolor{cvprblue}{rgb}{0.21,0.49,0.74}
\usepackage[pagebackref,breaklinks,colorlinks,allcolors=cvprblue]{hyperref}

\usepackage{amssymb}
\usepackage{pifont}
\newcommand{\cmark}{\ding{51}}%
\newcommand{\xmark}{\ding{55}}%

\def\paperID{50} 
\def\confName{CVPR}
\def\confYear{2025}

\title{Generating Synthetic Stereo Datasets using 3D Gaussian Splatting and Expert Knowledge Transfer}

\author{
Filip Slezák$^{1,2}$ \quad Magnus Kaufmann Gjerde$^{1}$ \quad Joakim Bruslund Haurum$^{1,3}$ \quad Ivan Nikolov$^{1}$\\ Morten S. Laursen$^{2}$ \quad Thomas B. Moeslund$^{1,3}$\\
$^{1}$Visual Analysis \& Perception Lab, Aalborg University\\ $^{2}$AGCO A/S, Denmark\quad $^{3}$Pioneer Centre for AI, Denmark
}

\begin{document}
\maketitle

\input{sec/0_abstract}    
\input{sec/1_intro}

\input{sec/2_related_work}

\input{sec/3_methodology}
\input{sec/4_experiments}

\input{sec/5_discussion_conclusion}

{
    \small
    \bibliographystyle{ieeenat_fullname}
    \bibliography{main}
}


\end{document}

%% file: sec/0_abstract.tex
\begin{abstract}
In this paper, we introduce a 3D Gaussian Splatting (3DGS)-based pipeline for stereo dataset generation, offering an efficient alternative to Neural Radiance Fields (NeRF)-based methods. To obtain useful geometry estimates, we explore utilizing the reconstructed geometry from the explicit 3D representations as well as depth estimates from the FoundationStereo model in an expert knowledge transfer setup. We find that when fine-tuning stereo models on 3DGS-generated datasets, we demonstrate competitive performance in zero-shot generalization benchmarks. When using the reconstructed geometry directly, we observe that it is often noisy and contains artifacts, which propagate noise to the trained model. In contrast, we find that the disparity estimates from FoundationStereo are cleaner and consequently result in a better performance on the zero-shot generalization benchmarks. Our method highlights the potential for low-cost, high-fidelity dataset creation and fast fine-tuning for deep stereo models. Moreover, we also reveal that while the latest Gaussian Splatting based methods have achieved superior performance on established benchmarks, their robustness falls short in challenging in-the-wild settings warranting further exploration.
\end{abstract}

%% file: sec/1_intro.tex
\section{Introduction}
\label{sec:intro}
Recovering the 3D structure of a scene captured from images is a widely researched problem that has exploded in popularity through the recent advances in monocular depth estimation as well as novel view synthesis approaches such as Neural Radiance Fields (NeRFs) \cite{Mildenhall2020NeRF} and Gaussian Splatting\cite{Kerbl20233DGS}.

Initial work in the field required the use of stereo image pairs and, through traditional algorithms like semi-global matching (SGM)\cite{hirschmuller2005accurate}, achieved remarkable performance using well-designed heuristics. This resulted in accurate disparity maps but still had several shortcomings due to the common challenges of stereo matching. These are, namely, textureless regions and occlusions - both of which make it very challenging to find reliable correspondences, leading to errors or a lack of prediction. 

With the introduction of large-scale datasets with accurate disparity labels, such as \cite{Mayer2015ALD}, deep learning-based disparity estimation methods were made possible. However, obtaining per-pixel labels is very challenging, typically addressed by using a simulated environment where accurate ground truth geometry can be directly extracted from the explicit scene representation. While the geometry labels are highly accurate, the realism gap between rendered and real-world images makes these methods perform suboptimal when applied on real-world samples.

Recent advancements include the addition of vision foundation models pre-trained for monocular depth estimation \cite{hu2024metric3d, yang2024depth} and powerful context networks, such as Dinov2 \cite{oquab2023dinov2}, which have greatly improved the networks' reasoning ability in the ill-posed regions. 
This motivated the use of monocular depth estimators such as DepthAnythingV2 \cite{yang2024depth} and Metric3Dv2 \cite{hu2024metric3d} to produce high-quality depth estimates of real images. Orthogonally, Tosi \etal \cite{Tosi2023NeRFSupervisedDS} instead simply collect data via mobile phone recordings and apply the recent advances in the field of Novel View Synthesis in order to obtain realistic images and cheap geometry labels. 
However, NeRFs, as used by \eg Tosi \etal, have a considerable shortcoming. The implicit representation of the scene fails to produce accurate, fully dense geometry, requiring a lot of filtering, which makes the disparity maps very sparse. As a consequence, a complicated training procedure requires photometric losses apart from the common L1 disparity loss to compensate for the lack of density. Such a training setup suffers from training instability, and the results are not replicable as outlined in greater detail in our experimentation section and observing the replication efforts of the academic community.

\begin{figure*}[t]
  \centering
  \includegraphics[width=\textwidth]{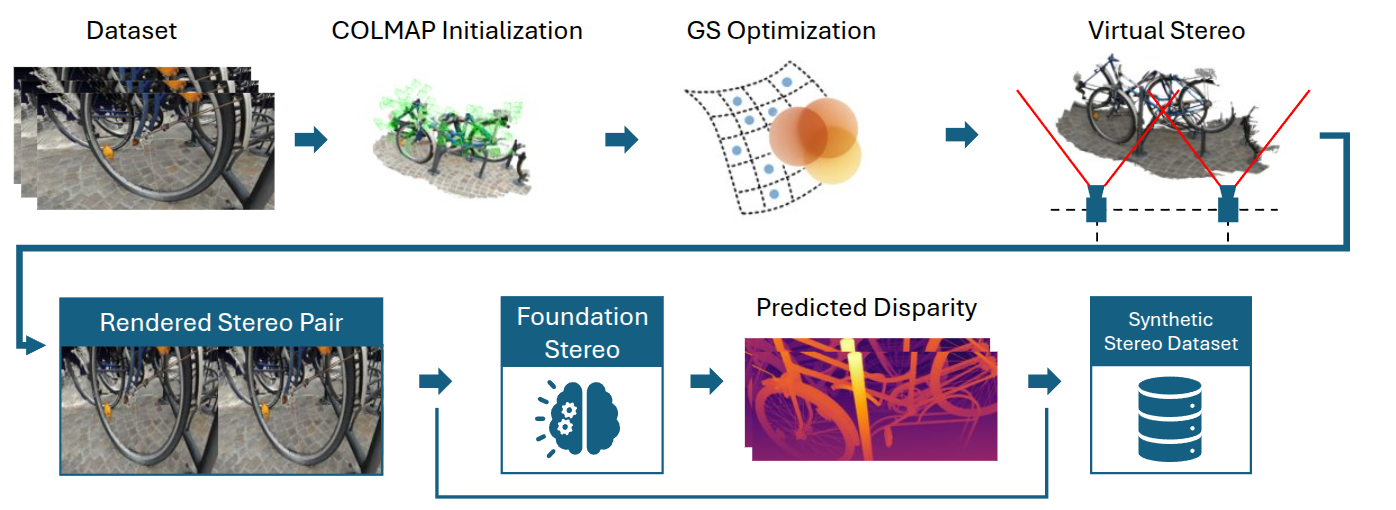}
  \caption{\textbf{Expert Knowledge Transfer using FoundationStereo.} Overview of the proposed expert knowledge transfer setup where stereo pairs rendered with Gaussian Splatting are supplied to FoundationStereo \cite{FoundationStereo}, which produces high-quality pseudo depth estimates. The process starts with the 3D Gaussian Splatting (3DGS) pipeline \cite{Kerbl20233DGS}, where COLMAP \cite{schoenberger2016mvs}\cite{schoenberger2016sfm} is used to initialize the optimization of 3DGS. Stereo image pairs are rendered after the scene has been fitted with 3DGS. As the underlying geometry of 3DGS is poorly reconstructed, the stereo pairs are instead sent to FoundationStereo, producing ground-truth disparity images. The generated synthetic stereo dataset is then used to train a lightweight stereo network, RAFT-Stereo \cite{lipson2021raft} (not depicted in this figure).}
  \label{fig:overview}
\end{figure*}

Instead, we investigate the newly introduced 3D Gaussian Splatting \cite{Kerbl20233DGS} based methods, which explicitly model the scene using Gaussians that can be converted to meshes \cite{Huang20242DGS, chen2024pgsr, Yu2024GaussianOF}. As these methods offer highly realistic images and seemingly accurate surface reconstruction, we hypothesize that they might be the missing link to the easy acquisition of stereo-matching datasets. Furthermore, our choice of 3DGS based methods is further motivated by their excellent rendering speed compared to NeRF based methods of comparable reconstruction quality. Concurrently, we are inspired by the recent development within monocular and stereo vision foundation models, and investigate whether using depth estimates from a large stereo vision foundation model \cite{FoundationStereo} outperforms the performance obtained with the reconstructed geometry, as shown in Figure~\ref{fig:overview}.

Using a fine-tuning training setup, we evaluate the effect of the different 3D Gaussian Splatting-based synthetic datasets, and compare directly with NeRF-Stereo. We do so by performing zero-shot evaluation on a set of previously unseen datasets commonly used within the stereo depth estimation field. 

Our contributions are as follows:
\begin{itemize}
    \itemsep0em
    \item We qualitatively and empirically find that the rendered disparity from 3DGS and the reconstructed mesh of PGSR are insufficient for fine-tuning a RAFT-Stereo network.
    \item We propose a knowledge transfer setup where predicted depth estimates from FoundationStereo are used as ground truth for 3DGS stereo renderings.
    \item We find that fine-tuning a RAFT-Stereo network using our proposed expert knowledge transfer setup consistently outperforms other methods, including the state-of-the-art NeRF-Stereo method.
\end{itemize}

%% file: sec/2_related_work.tex
\section{Related Works}\label{sec:related_work}

\subsection{Disparity Estimation Algorithms}
Large-scale disparity map estimation became possible with the development of SGM algorithms \cite{hirschmuller2005accurate}, which optimized the calculated disparity along preset directional paths, instead of purely global optimization, which can be costly, or local methods that tended to leave artifacts around edges of objects. With the rise of deep learning models, architectures like PSM-Net \cite{chang2018pyramid} and models built on top of it \cite{huang2021stereo}, using features extracted by ResNet and a pyramidal structure which utilizes features from different scales. Later on, the work presented in RAFT-Stereo \cite{teed2021raft} and later on in RAFT-3D \cite{lipson2021raft} models the disparity and full 3D scene flow between consecutive frames and iteratively optimizes them using GRU layers. Taking the iterative optimization idea and expanding on it through the use of Geometry Encoding Volume is the IGEV-Stereo \cite{xu2023iterative}. It incorporates more local details through the iterative optimization of the disparity maps to capture smaller 3D surface features. One problem that persists in both RAFT and IGEV models is the ambiguity in smooth and featureless regions, which translates into incorrect disparity values propagated through the iterative optimization. The Selective-Stereo model \cite{wang2024selective}, which introduces the Selective Recurrent Unit (SRU), instead of the normally used GRU ones, together with a Contextual Spatial Attention (CSA) module, tries to mitigate this problem. It aggregates disparity information at different levels of detail and frequencies to capture hidden disparity information, especially in edge regions and at featureless areas. As transformers have proven to provide high-quality and robust results in many other computer vision fields, their use in disparity estimation has also proven extremely useful. Both the transformer self-attention and relative pixel distance encoding of STTR \cite{li2021revisiting} and the positional embeddings and pre-training of CroCoV2 \cite{weinzaepfel2023croco} boost their accuracy compared to CNN-based models and can handle large discrepancies in the estimated disparity.


\subsection{Stereo Datasets and Training Paradigm}
The advent of deep learning-based disparity estimation methods can be traced back to the origin of large-scale datasets that contain accurate disparity labels, such as the seminal work by Mayer et al. \cite{Mayer2015ALD}. However, obtaining per-pixel labels is very challenging, typically addressed by using a simulated environment where accurate ground truth geometry can be directly extracted from the synthetic rendering pipeline. However, while the geometry labels are highly accurate, the rendered images have a realism gap compared to images captured in the real world. While realistic images can be easily captured by building a real stereo pair and capturing the real world, extracting accurate geometry becomes the primary challenge. Accurate dense geometry can be captured by either using structured light projectors \cite{Scharstein2014HighResolutionSD} or using an expensive Laser Scanner \cite{Schps2017AMS}. Both of these methods require significant post-processing and static scenes, but they achieve a dense geometry reconstruction. An additional approach is to use a multi-beam LiDAR, which can also support dynamic scenes, such as \cite{Gehrig2021DSECAS} \cite{Geiger2012AreWR} and \cite{Geiger2013VisionMR}. Still, these require post-processing to address the typical scanning distortion, accurate calibration, and offer only sparse depth maps due to the current LiDAR sensor limitations. Furthermore, neither synthetic nor real stereo dataset collection methodology offers flexibility to easily capture custom scenes to improve disparity estimation performance on specific tasks. Both methodologies are prohibitively expensive, as custom synthetic scenes require experts to create custom digital assets, while the real dataset requires a sensor suite with state-of-the-art depth sensors. Recently, the advancements in novel view synthesis (NVS) have opened up the possibility to use freely captured images to generate training data for stereo depth estimation \cite{Tosi2023NeRFSupervisedDS, FilipCVPRW, SelfEvlovling3DGS} and optical flow \cite{ADFactory}. However, due to the limitations of NeRFs \cite{Mildenhall2020NeRF}, it is not possible to extract accurate geometry, and the training procedure requires a complicated loss function to achieve good performance, deviating from the simple L1 loss used in the state-of-the-art works, which use existing stereo datasets.

\definecolor{gold}{rgb}{1.0, 0.84, 0.0}
\definecolor{silver}{rgb}{0.75, 0.75, 0.75}
\definecolor{bronze}{rgb}{0.8, 0.5, 0.2}

\subsection{Novel View Synthesis and Mesh Extraction}
Given a freely captured set of RGB images, novel view synthesis encodes the scene in a representation which can be queried to obtain novel views. Pioneering work NeRFs \cite{Mildenhall2020NeRF} has revolutionized the field, achieving unprecedented realism. Subsequent works have improved on the original idea \cite{Barron2021MipNeRFAM}, offering improved visuals, but at the cost of rendering time. While some works have also explored improving the rendering speeds \cite{Barron2023ZipNeRFAG}, it came at the cost of visual quality. The trade-off between rendering speed and visual quality has been addressed in the work by Kerbl et al. \cite{Kerbl20233DGS}, offering superior visual fidelity while achieving real-time rendering speeds. Furthermore, unlike NeRFs, which embed the scene implicitly in the weights of the neural network, 3D Gaussian Splatting presents an explicit geometric representation of the scene. Despite this, the original formulation is not suitable for accurate geometry extraction, presenting surface artifacts. The most recent works tackle this issue by structuring the scene composition in a manner that allows them to extract accurate meshes. For example, 2DGS  \cite{Huang20242DGS} employs a direct depth loss to the Gaussian primitives, and collapses one of the splat axes, representing the scene as a composition of 2D discs which are aligned with the surfaces, offering a better starting point for geometry extraction. Gaussian opacity fields \cite{Yu2024GaussianOF} aim to remove the extraction of the meshes as a post-processing step and extract accurate surfaces directly from the Gaussian splat cloud. However, all of these methods use only the information in the images, which can be problematic for textureless areas where multiple 3D geometries can render to the same 2D images. To correct this problem, researchers are looking to include pretrained networks, which can be a powerful regularizer for the ill-posed textureless regions of the image. Gaussian Surfels \cite{Dai2024HighqualitySR} uses a monocular normal estimation based on \cite{Eftekhar2021OmnidataAS}.

%% file: sec/3_methodology.tex
\begin{figure*}[htbp]
    \centering
    \begin{tabular}{@{}c@{}c@{}c@{}c@{}}
            \multicolumn{4}{c}{\textbf{Mesh reconstruction comparison applied to the NS dataset \cite{Tosi2023NeRFSupervisedDS}}}  \\ 

             Dataset image & PGSR & 2DGS & GOF \\

            \includegraphics[width=0.2\textwidth]{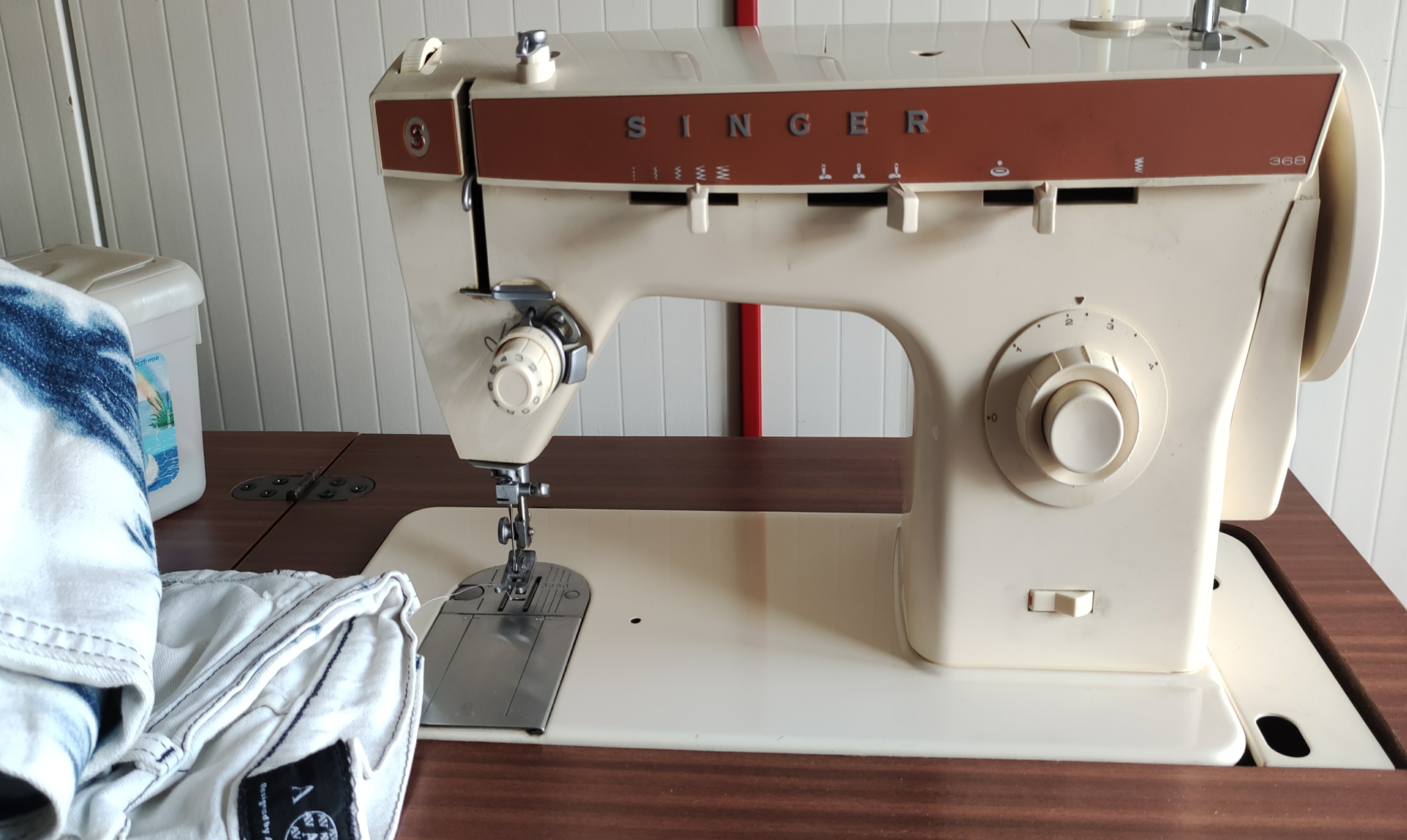} & 
            \includegraphics[width=0.2\textwidth]{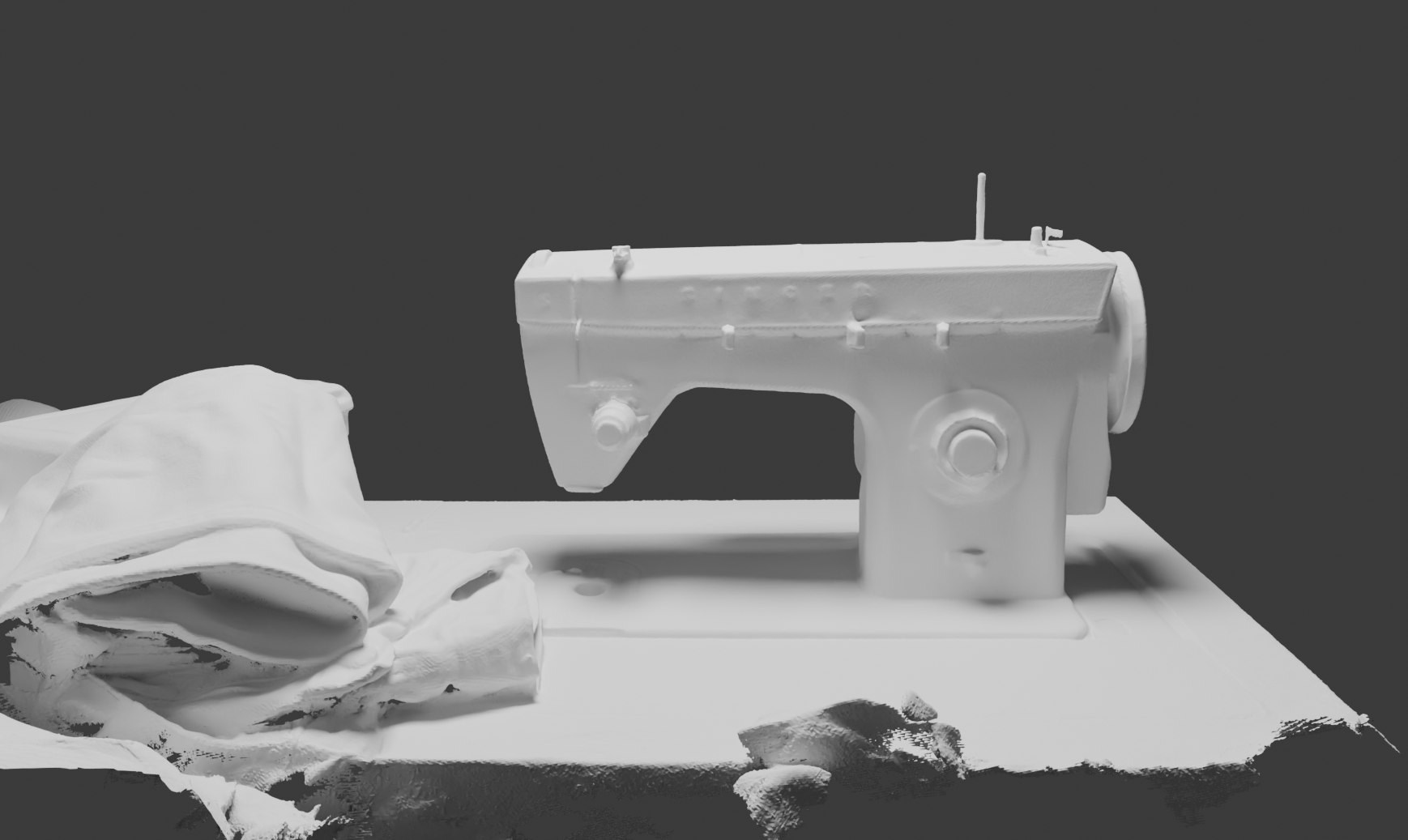} & 
            \includegraphics[width=0.2\textwidth]{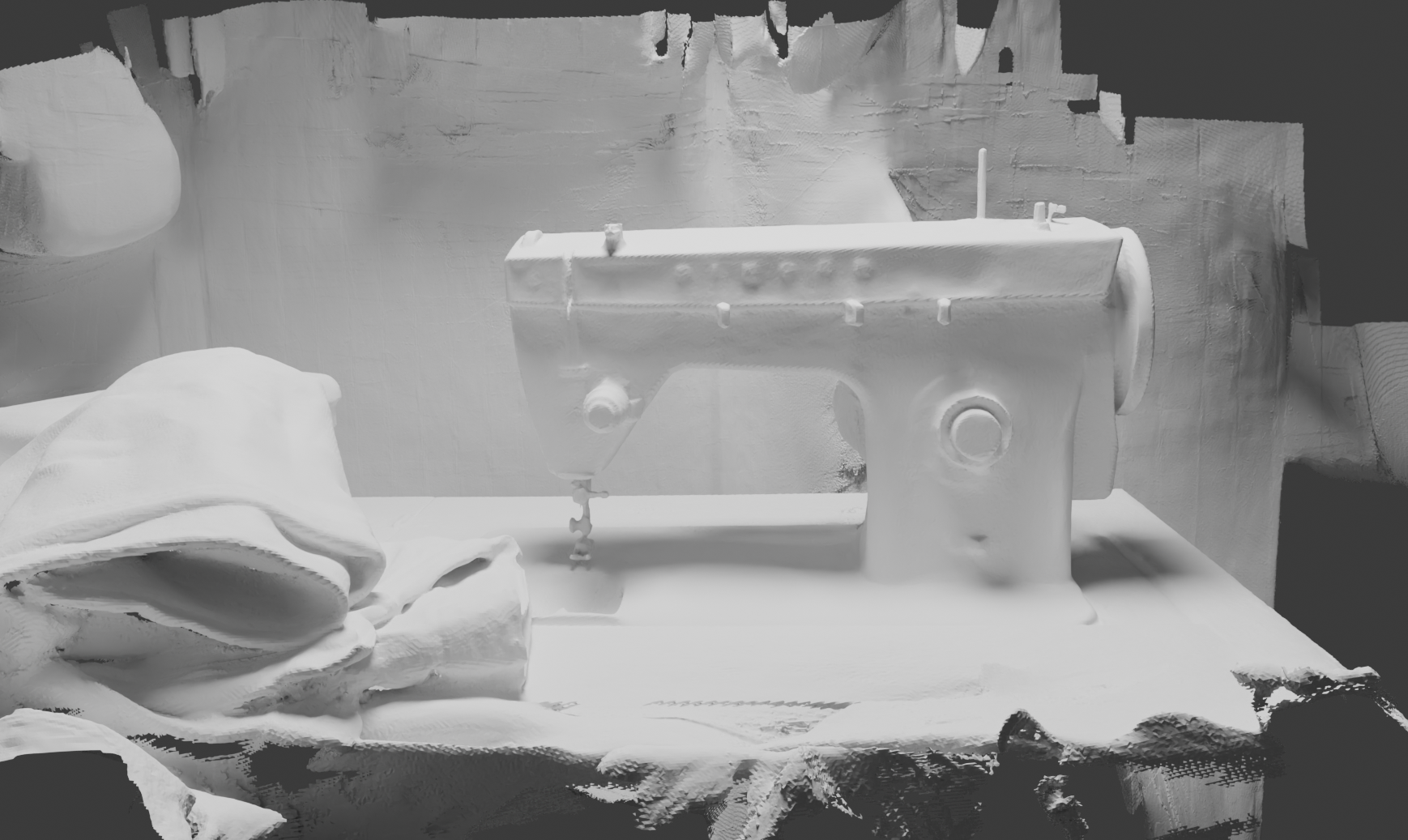} & 
            \includegraphics[width=0.2\textwidth]{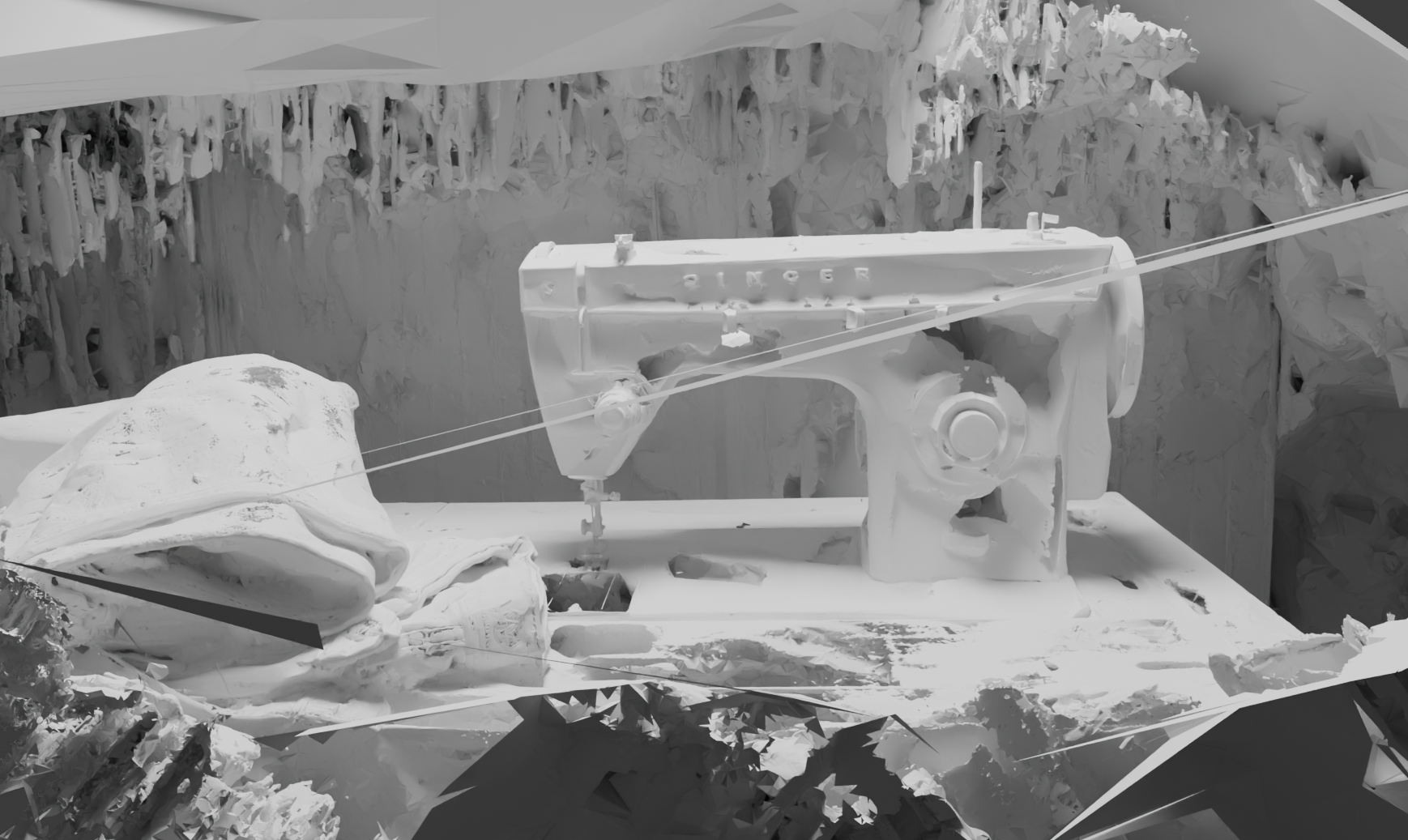} \\ 

            \includegraphics[width=0.2\textwidth]{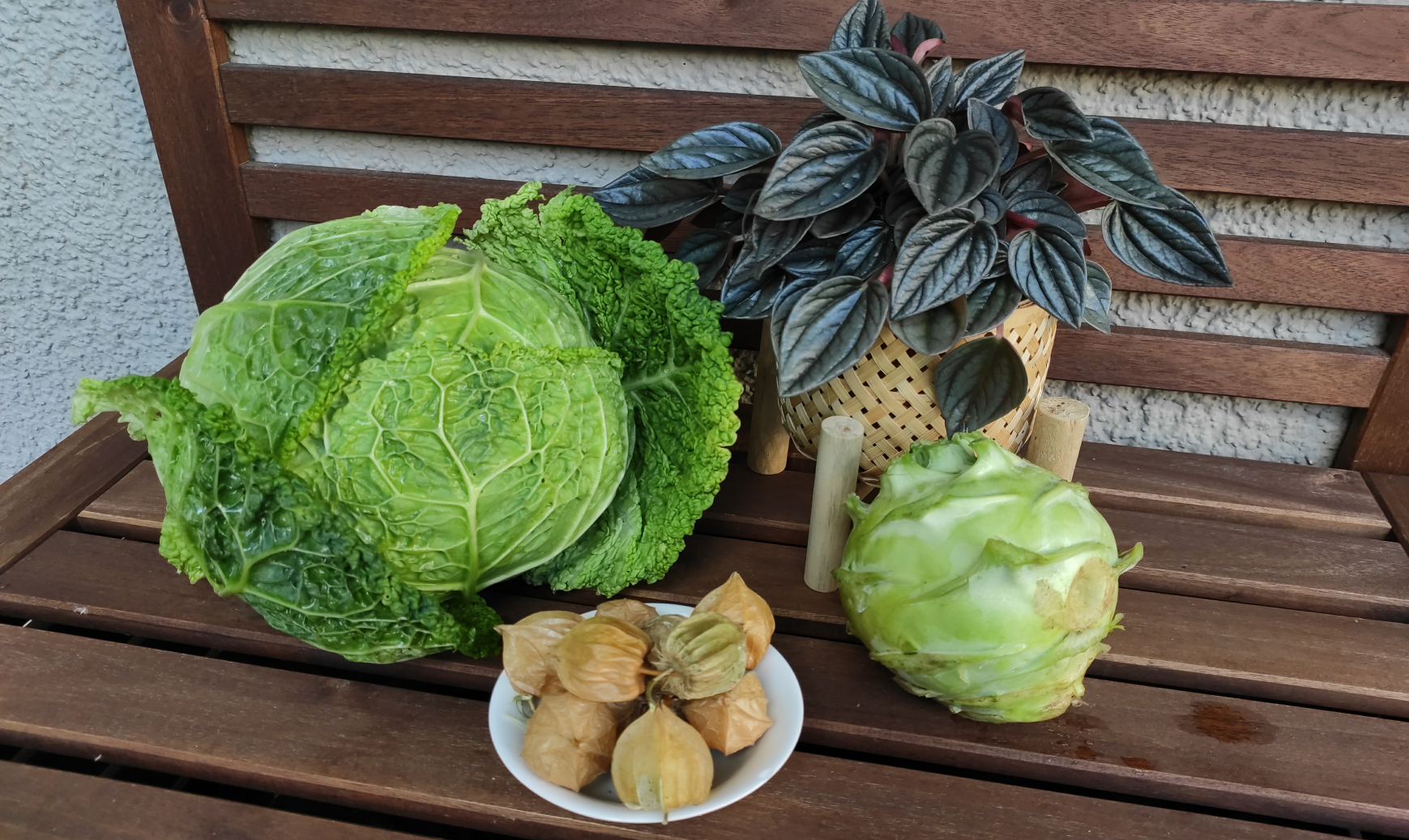} & 
            \includegraphics[width=0.2\textwidth]{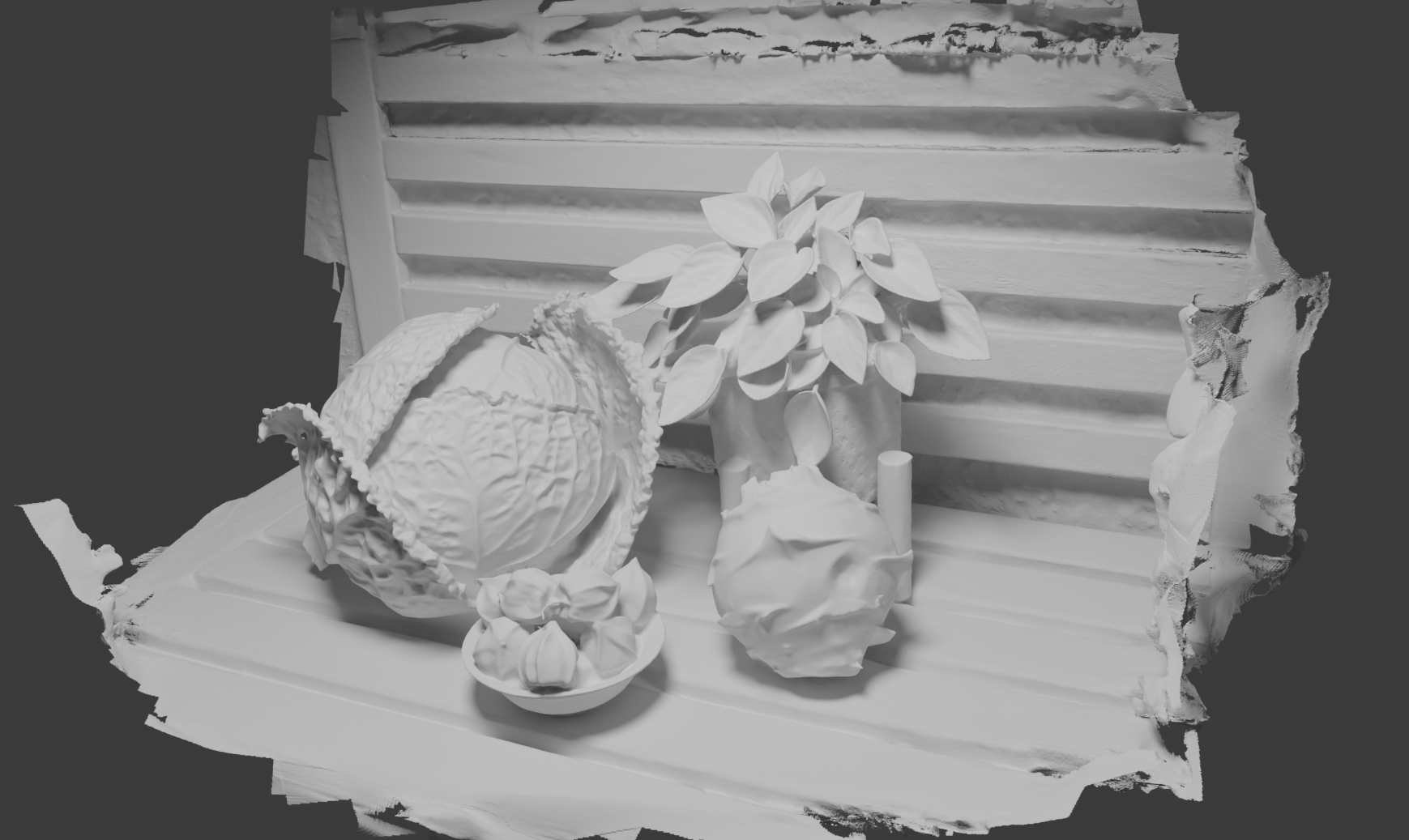} & 
            \includegraphics[width=0.2\textwidth]{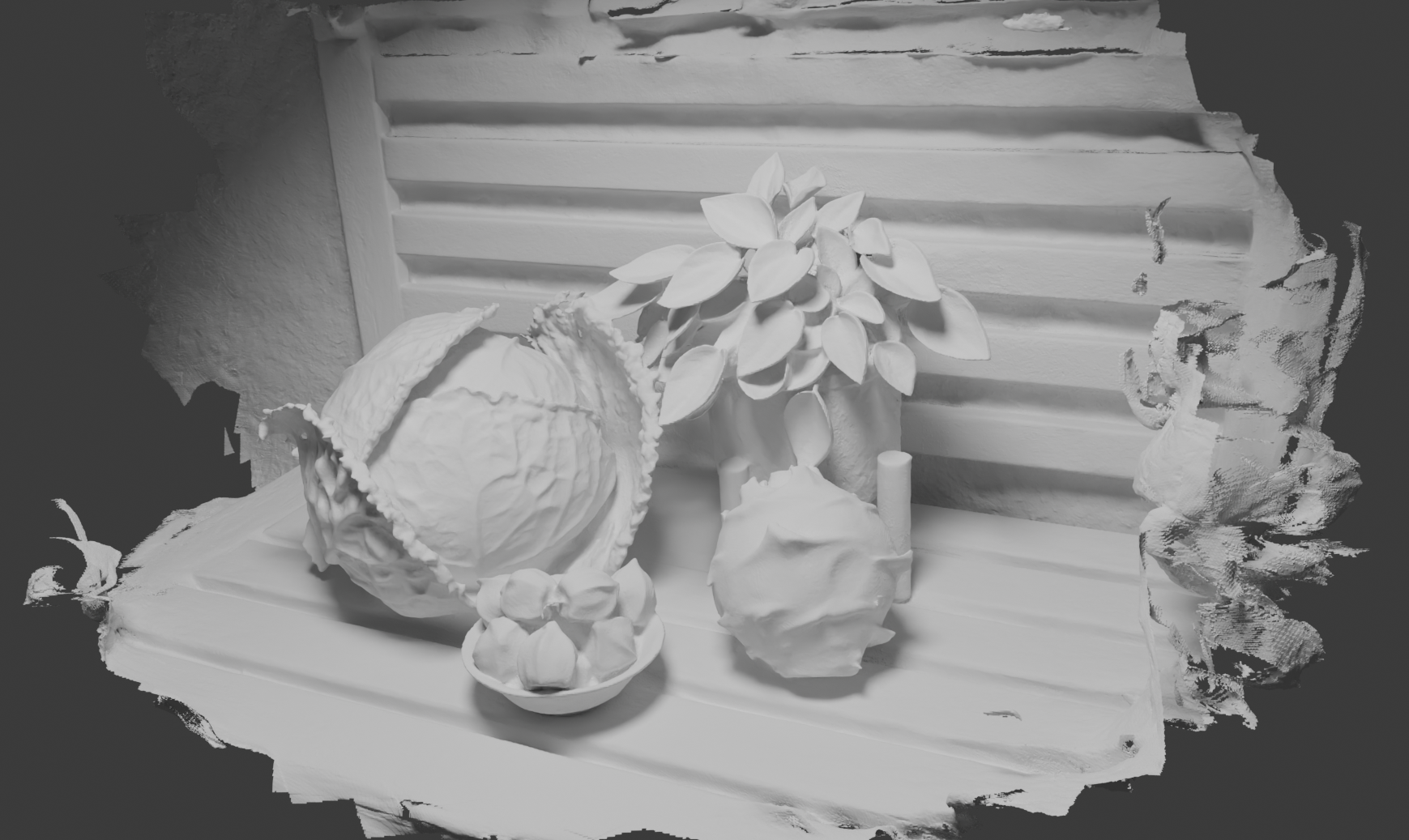} & 
            \includegraphics[width=0.2\textwidth]{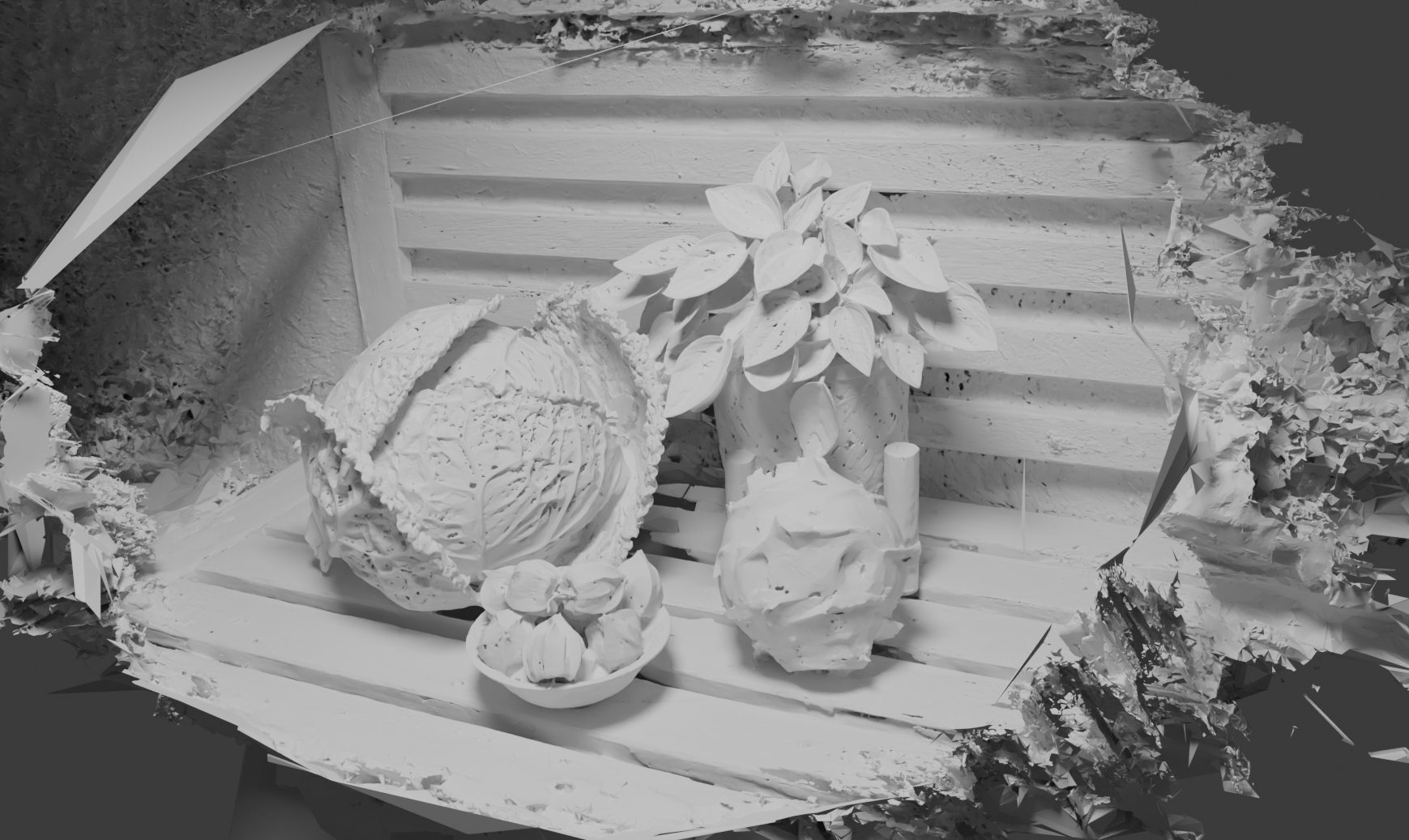} \\  
            
            \includegraphics[width=0.2\textwidth]{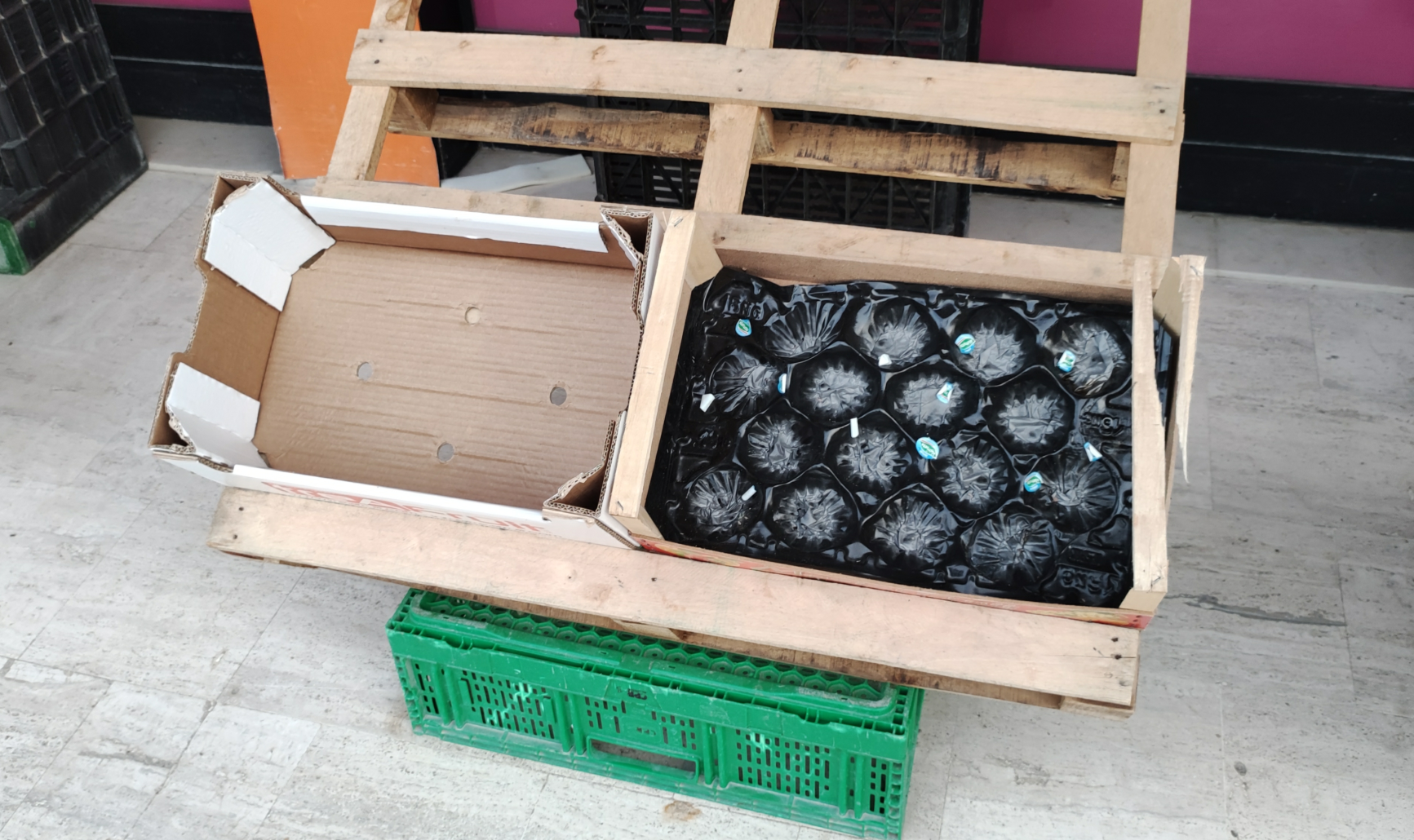} & 
            \includegraphics[width=0.2\textwidth]{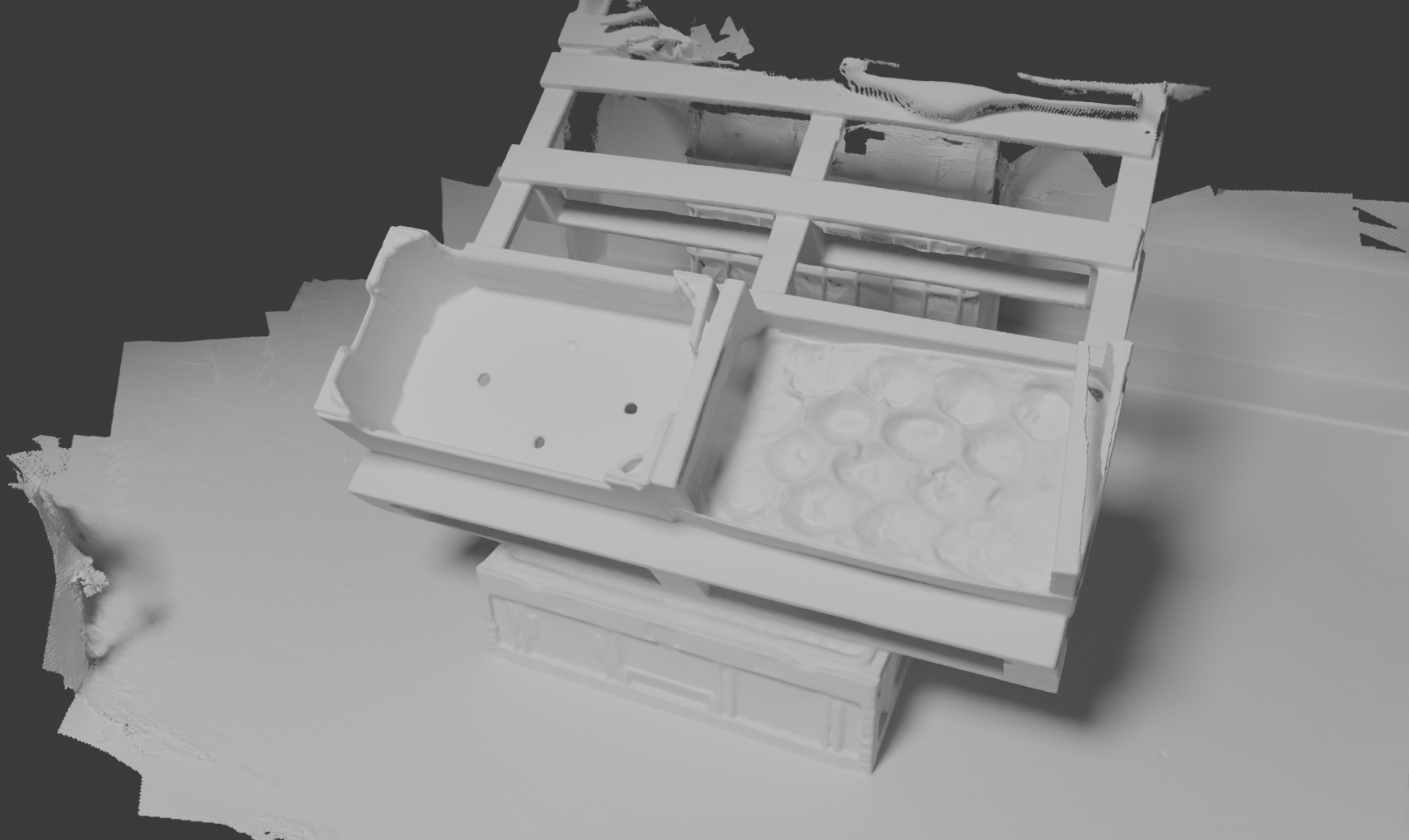} & 
            \includegraphics[width=0.2\textwidth]{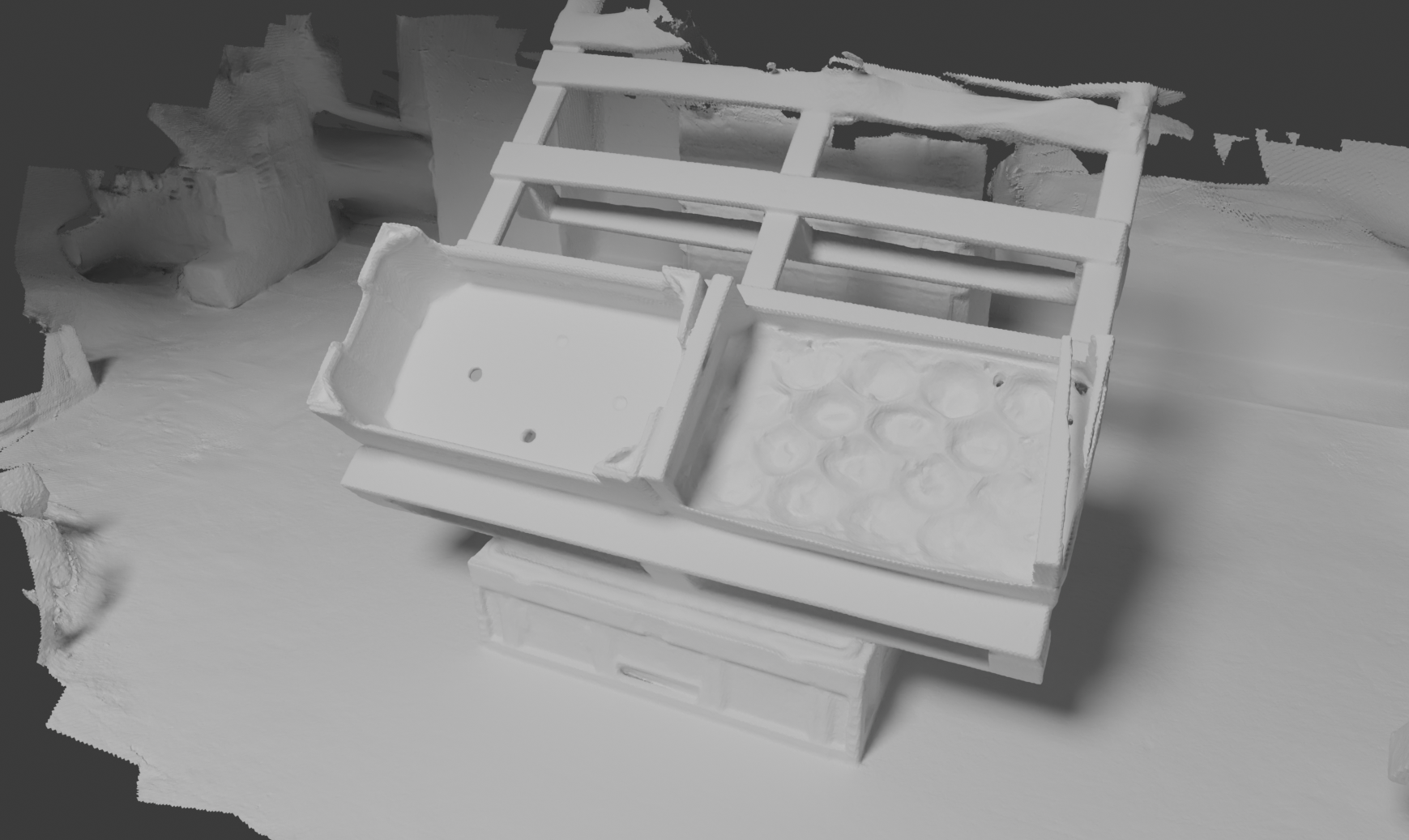} & 
            \includegraphics[width=0.2\textwidth]{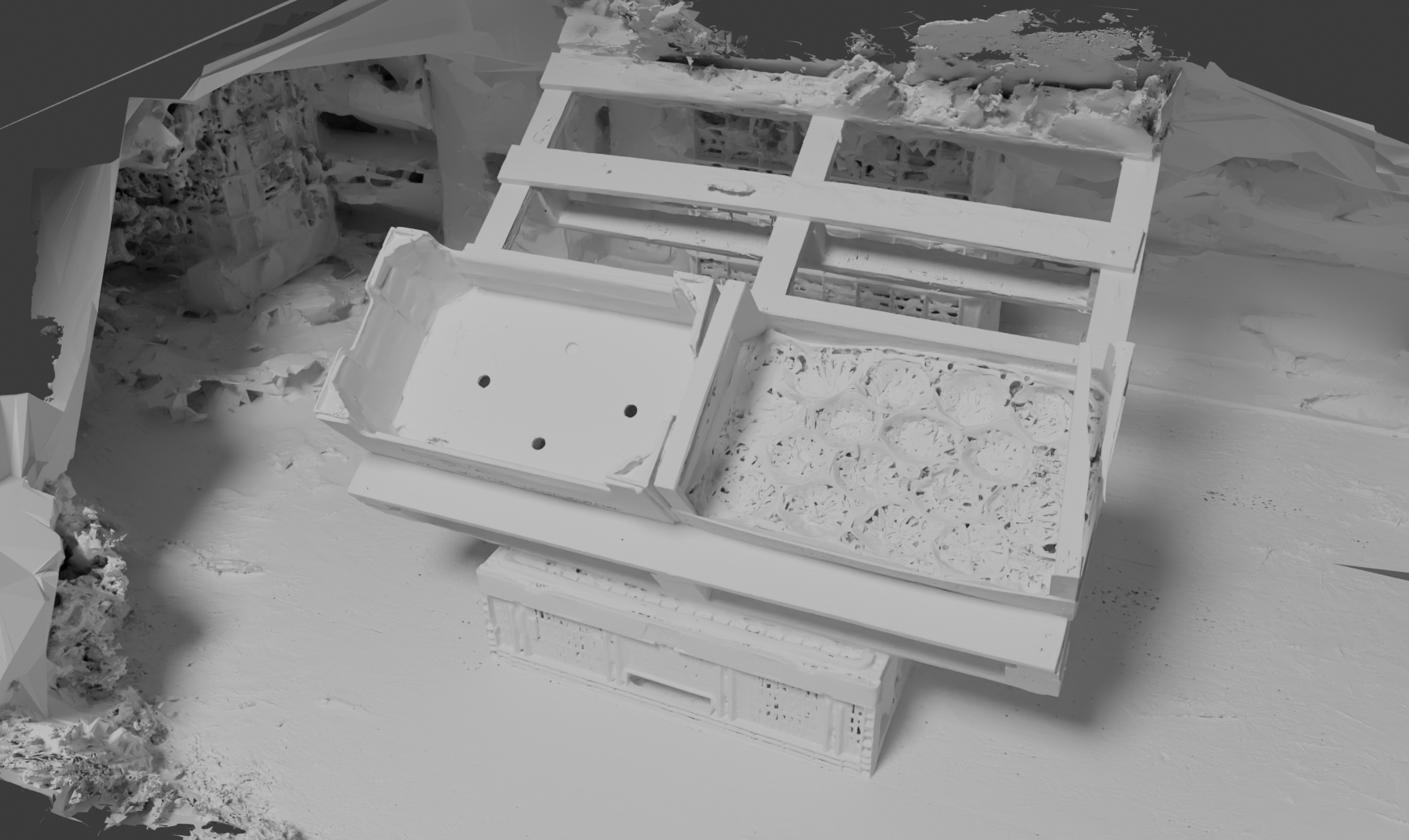} \\ 
            
            \multicolumn{4}{c}{\textbf{Rendered and backprojected disparity image}}  \\

            3DGS & PGSR & 2DGS & GOF \\

            \includegraphics[width=0.2\textwidth]{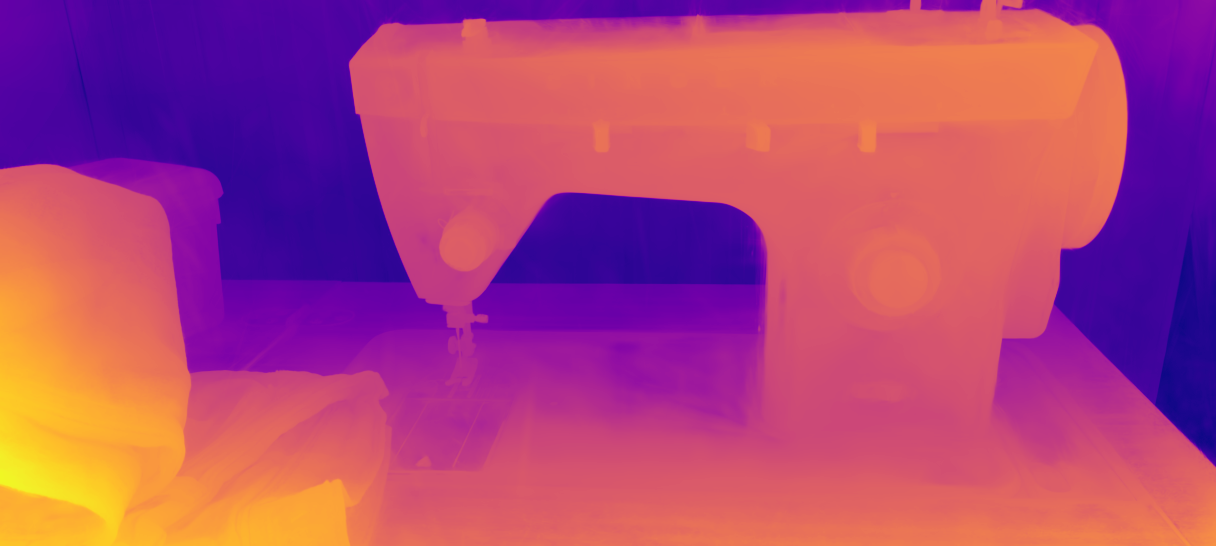} & 
            \includegraphics[width=0.2\textwidth]{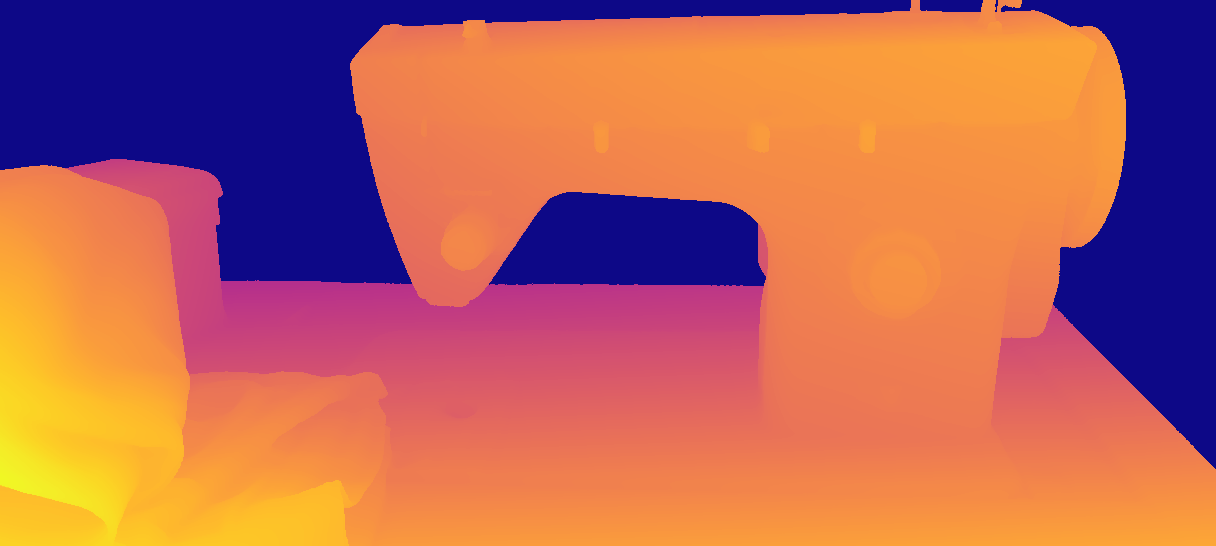} & 
            \includegraphics[width=0.2\textwidth]{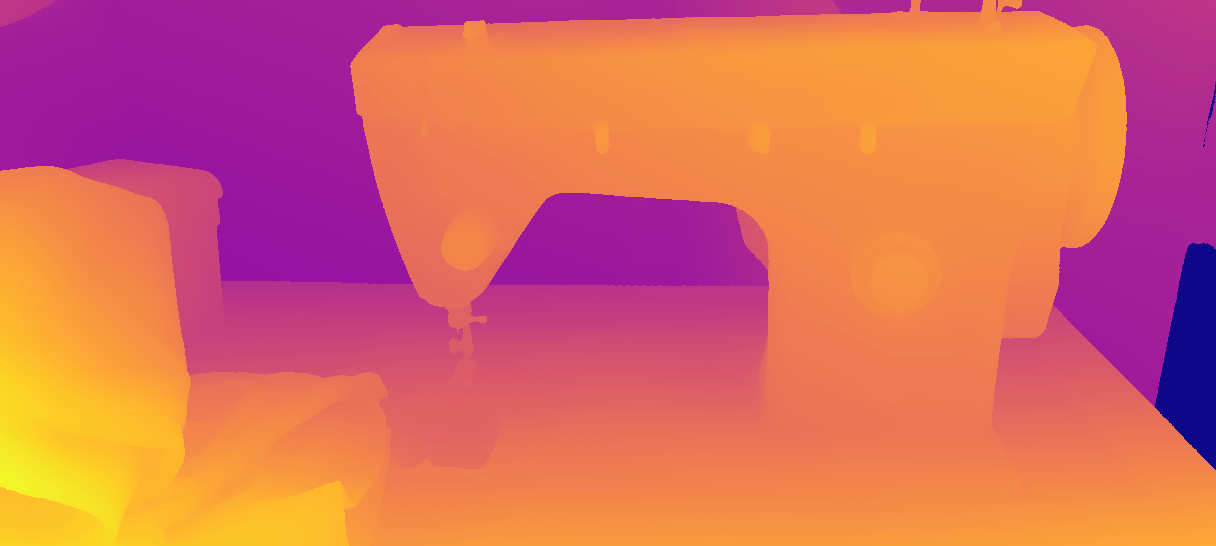} & 
            \includegraphics[width=0.2\textwidth]{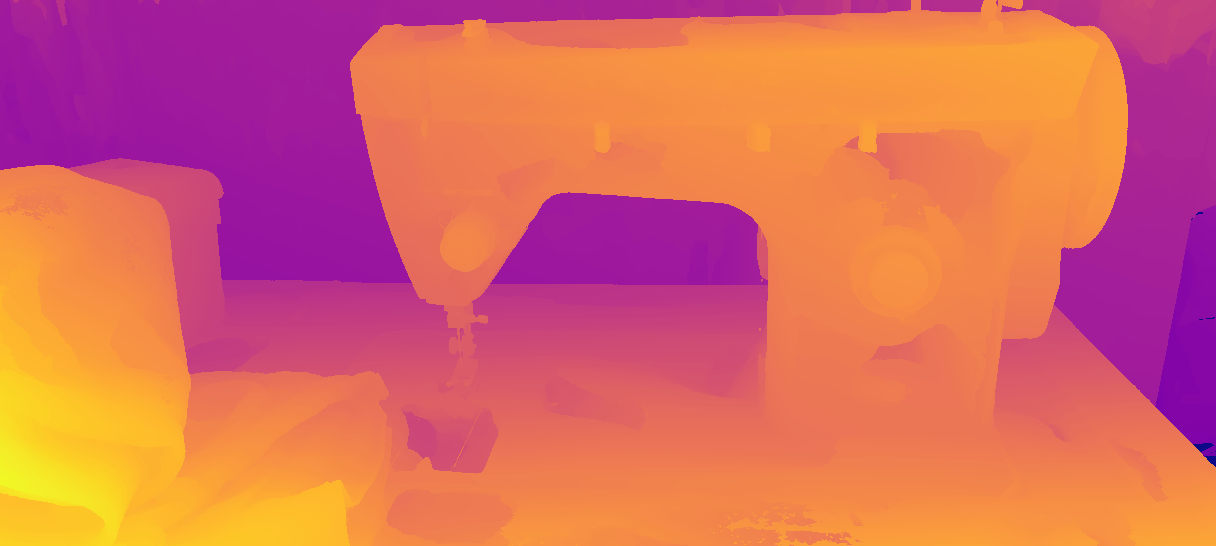} \\
                                        
            \includegraphics[width=0.2\textwidth]{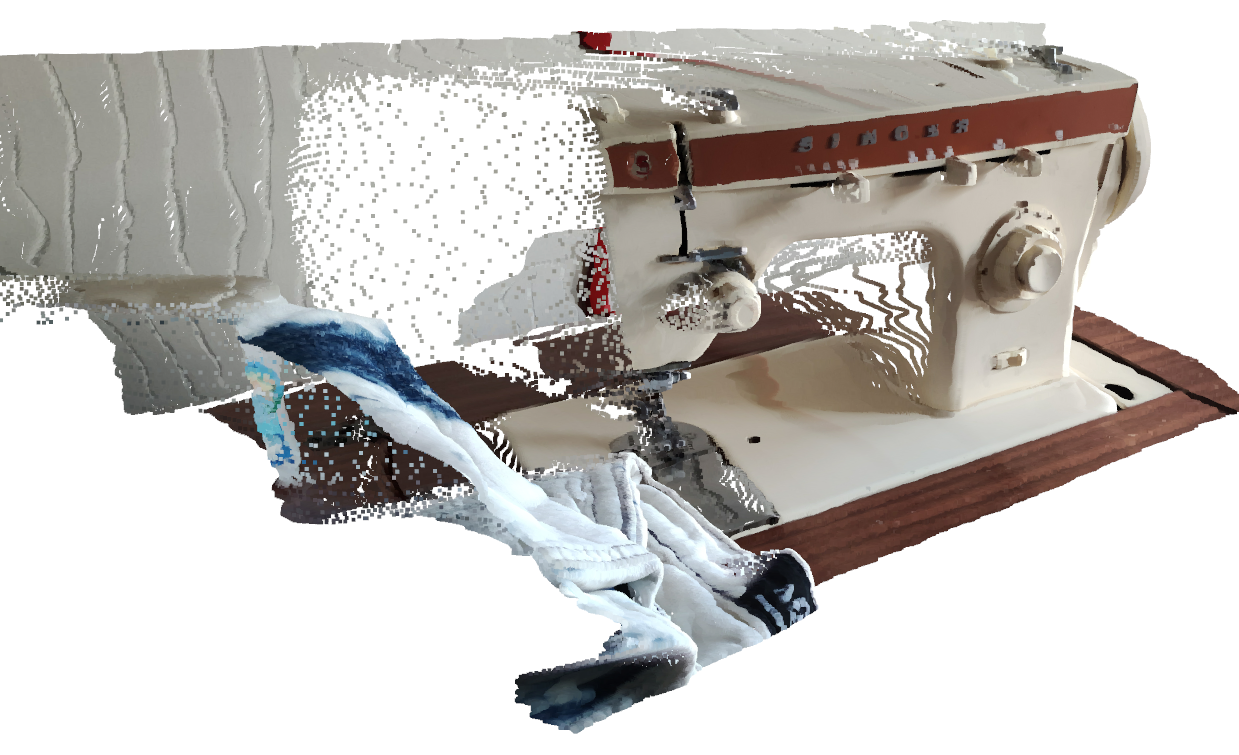} & 
            \includegraphics[width=0.2\textwidth]{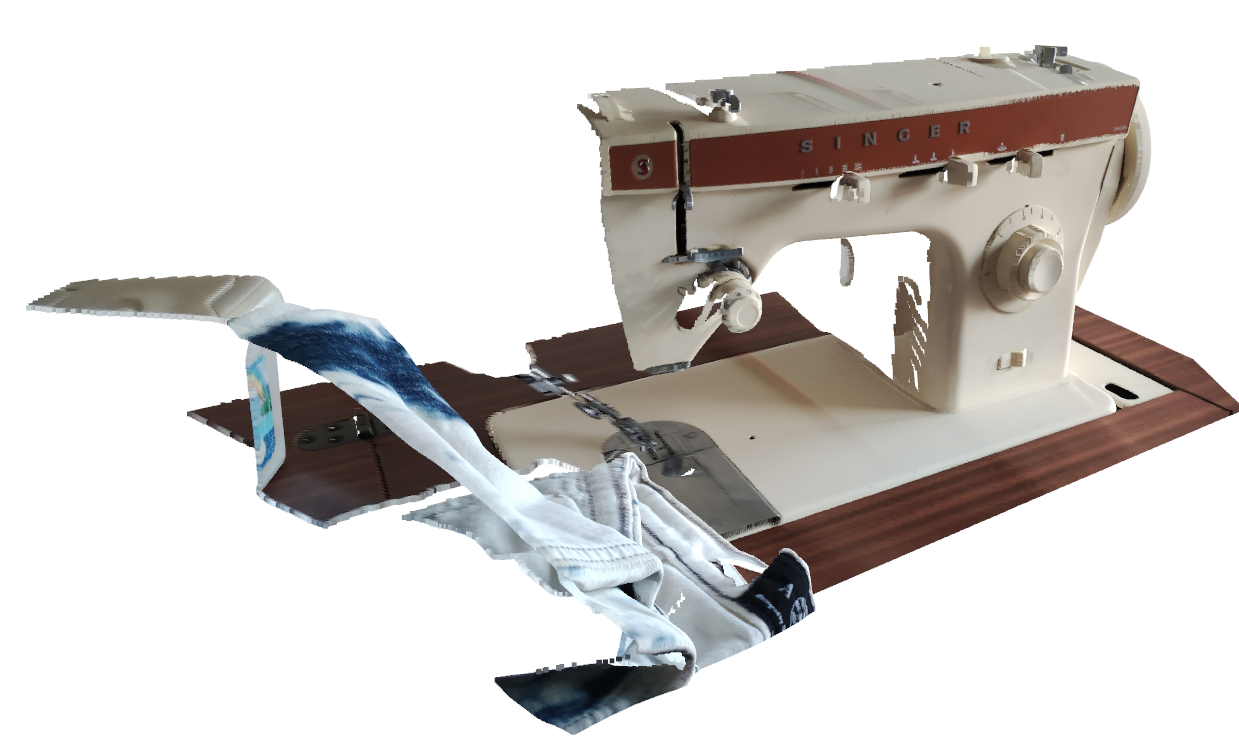} & 
            \includegraphics[width=0.2\textwidth]{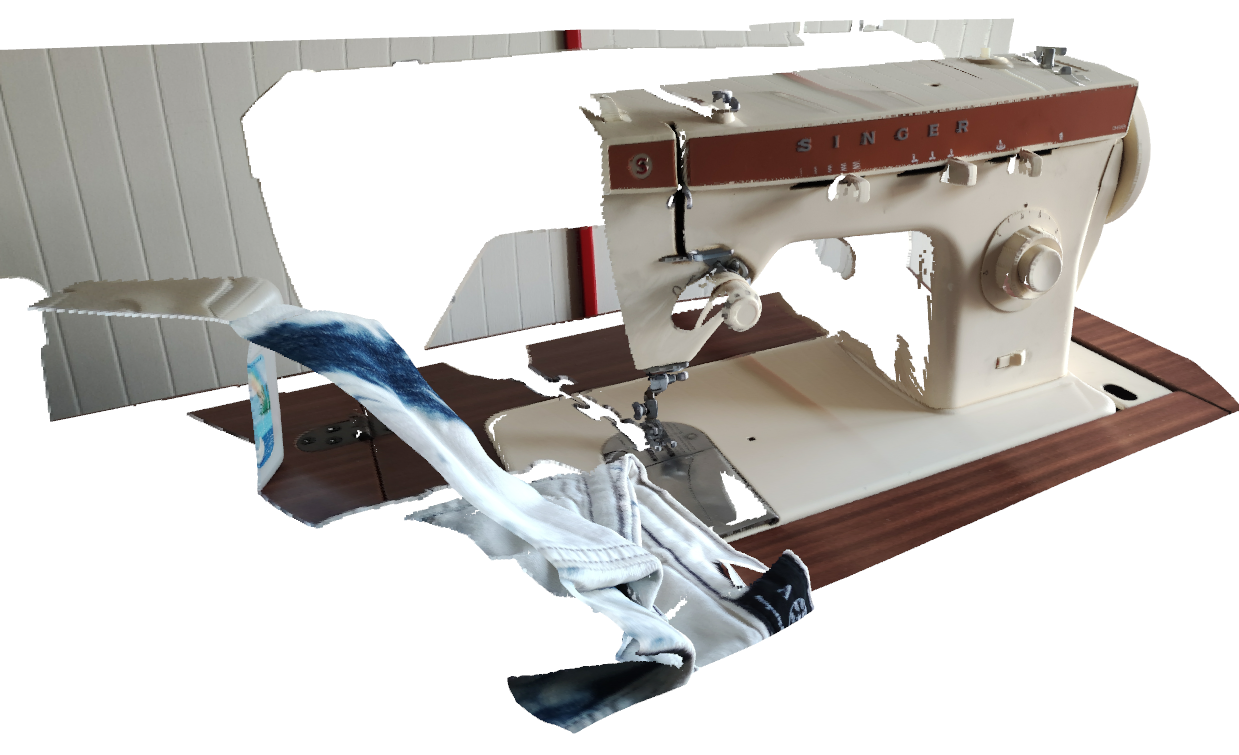} & 
            \includegraphics[width=0.2\textwidth]{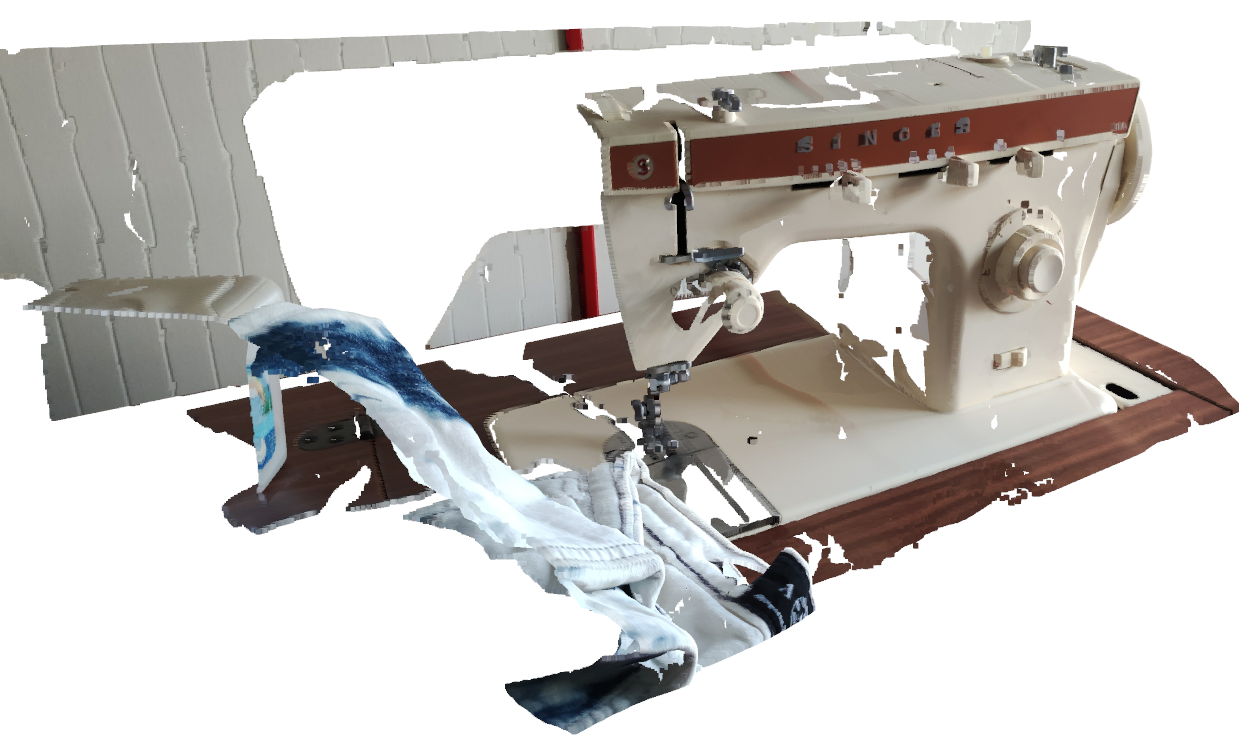} \\

            \includegraphics[width=0.2\textwidth]{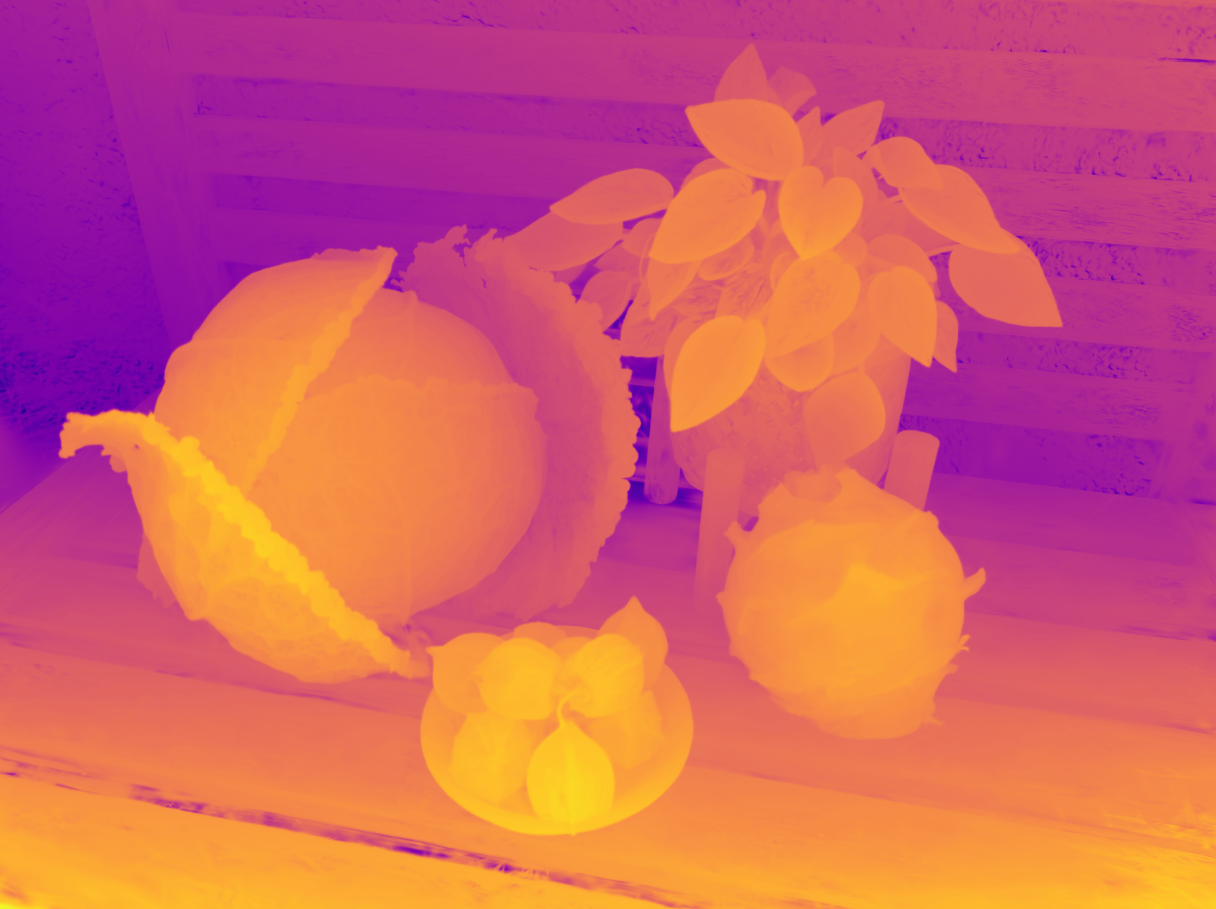} & 
            \includegraphics[width=0.2\textwidth]{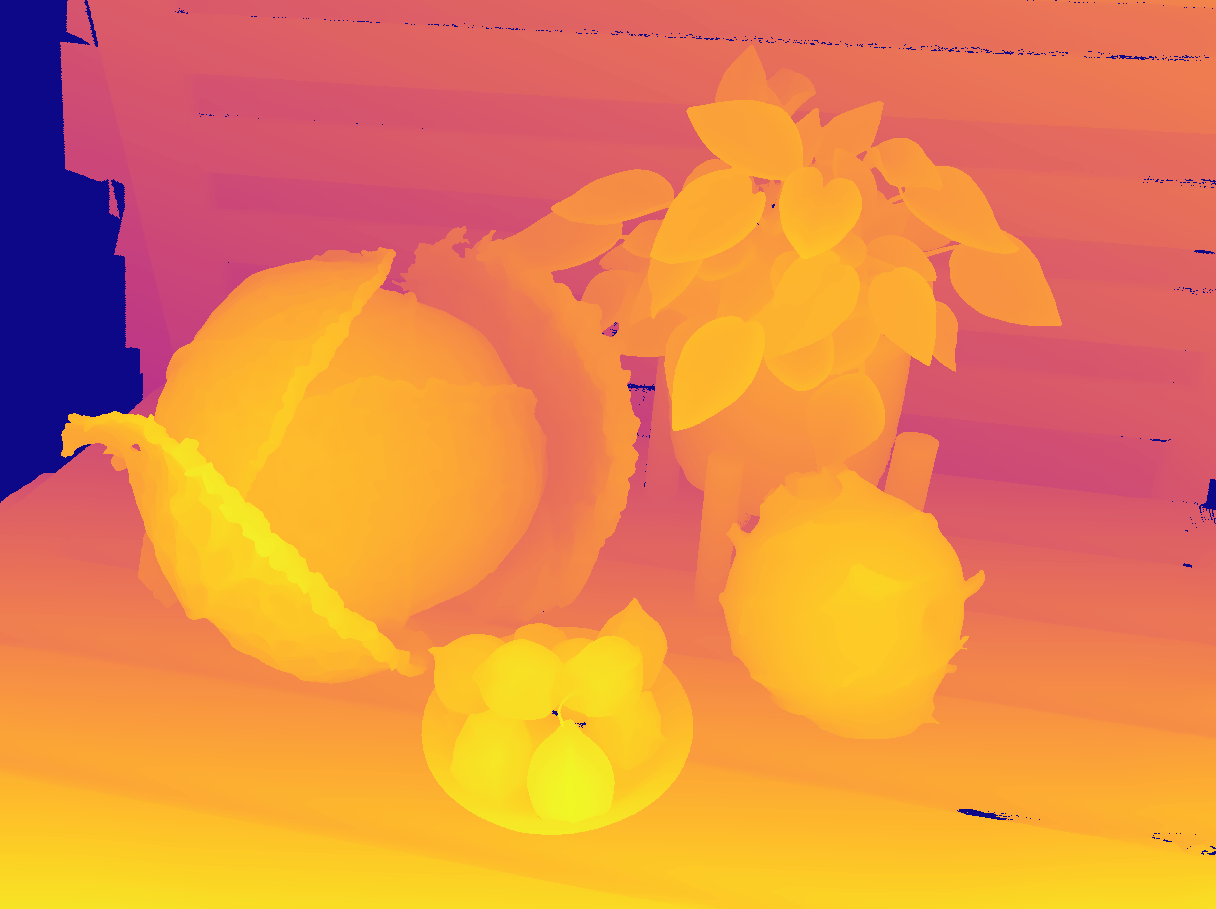} & 
            \includegraphics[width=0.2\textwidth]{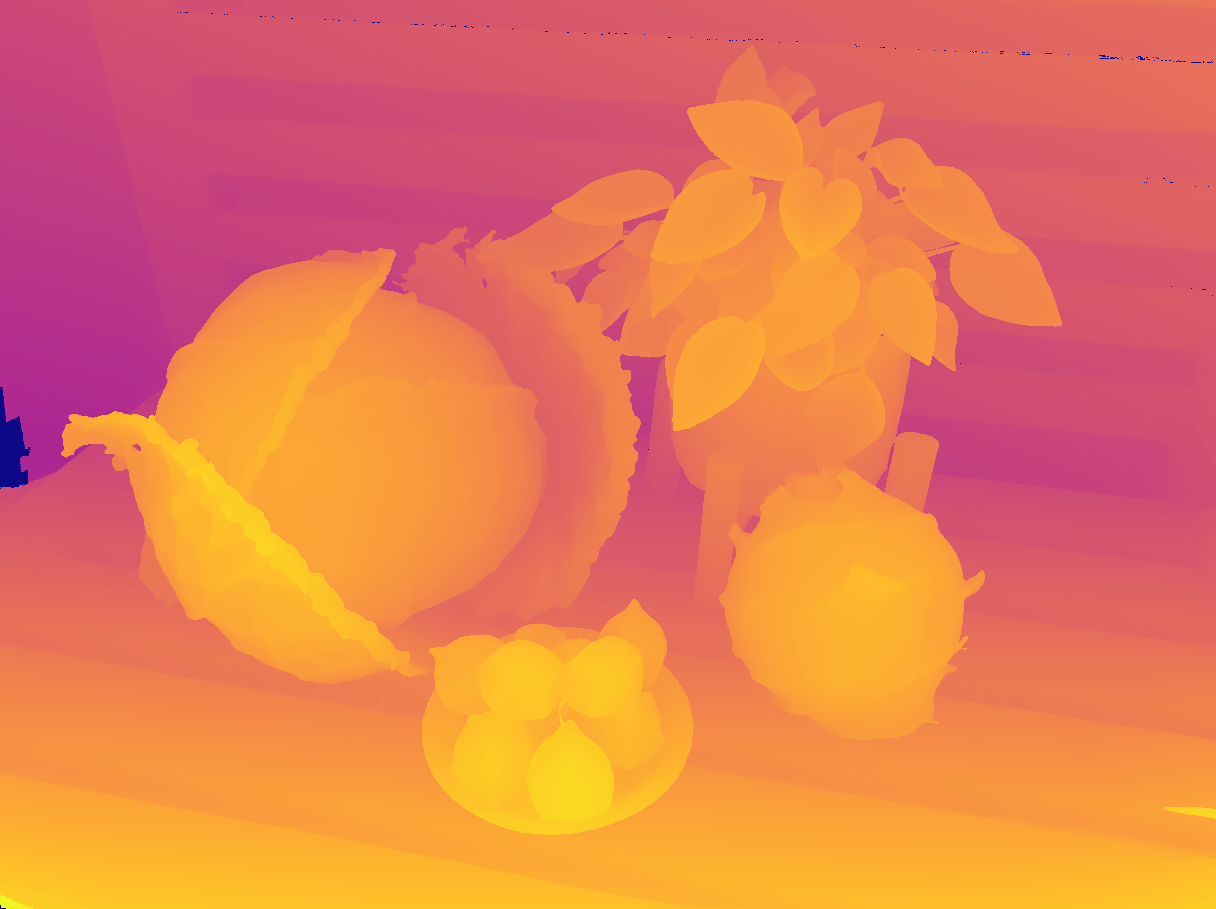} & 
            \includegraphics[width=0.2\textwidth]{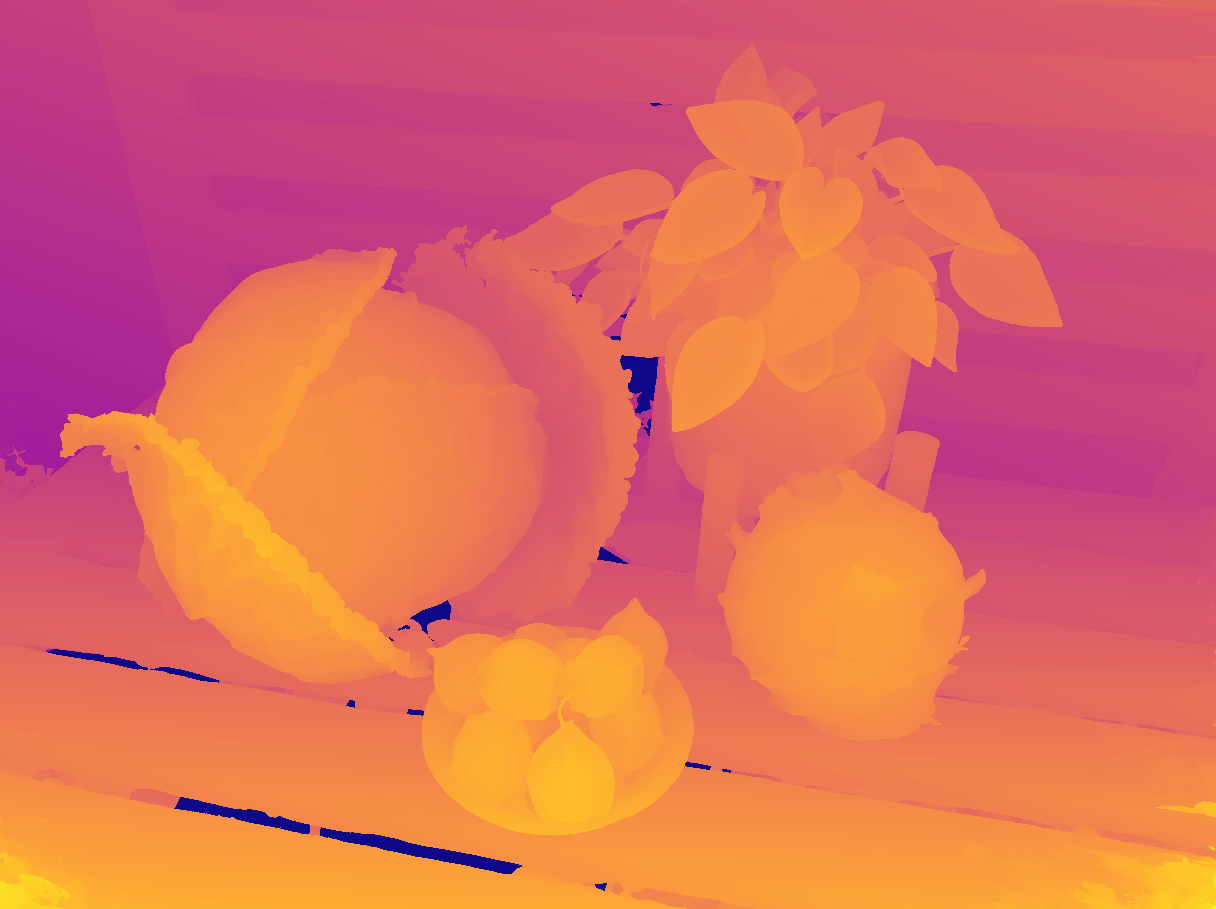} \\
                                        
            \includegraphics[width=0.2\textwidth]{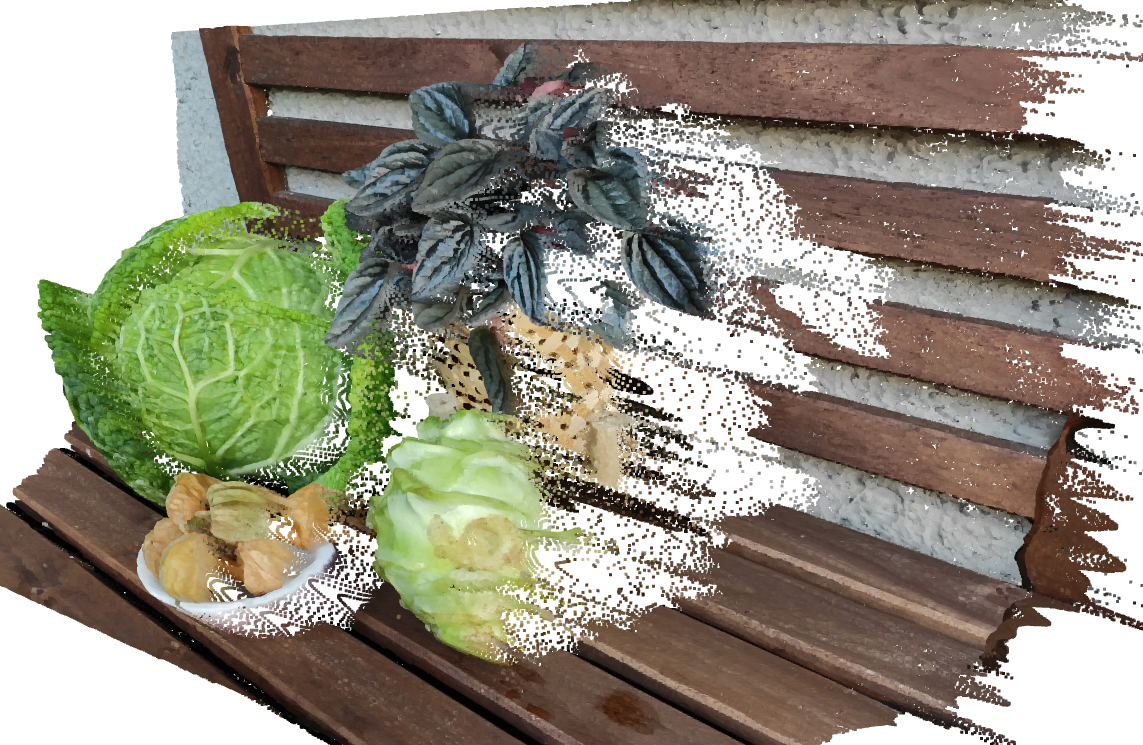} & 
            \includegraphics[width=0.2\textwidth]{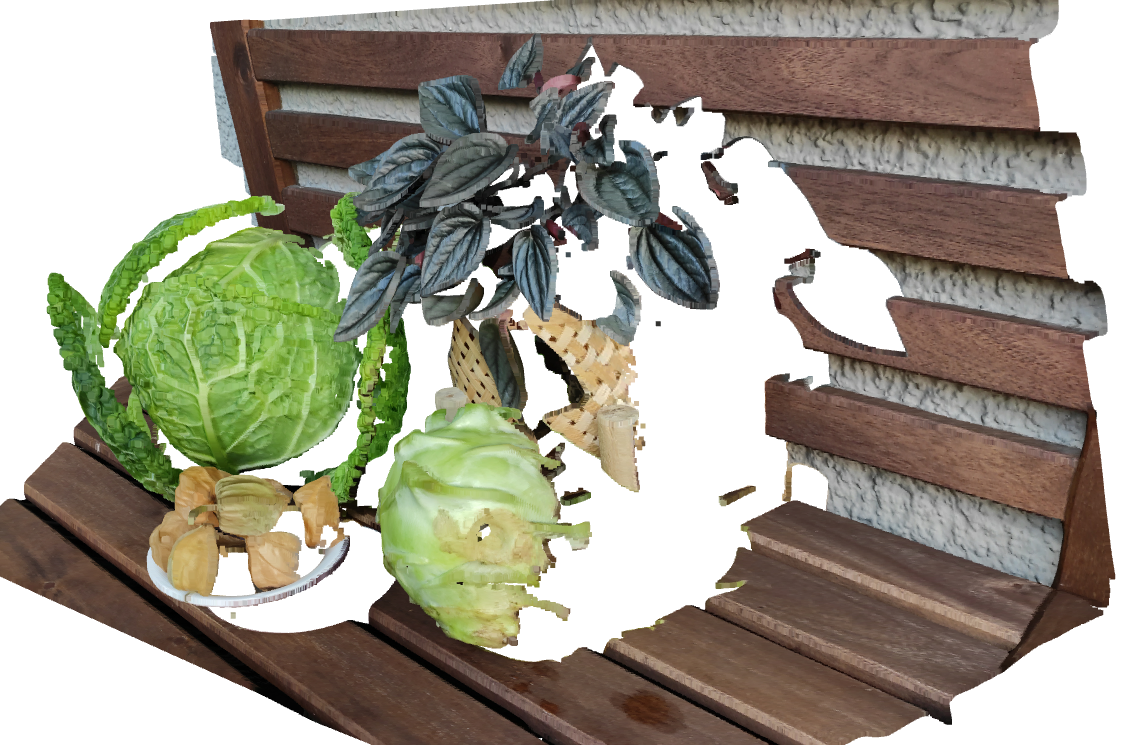} & 
            \includegraphics[width=0.2\textwidth]{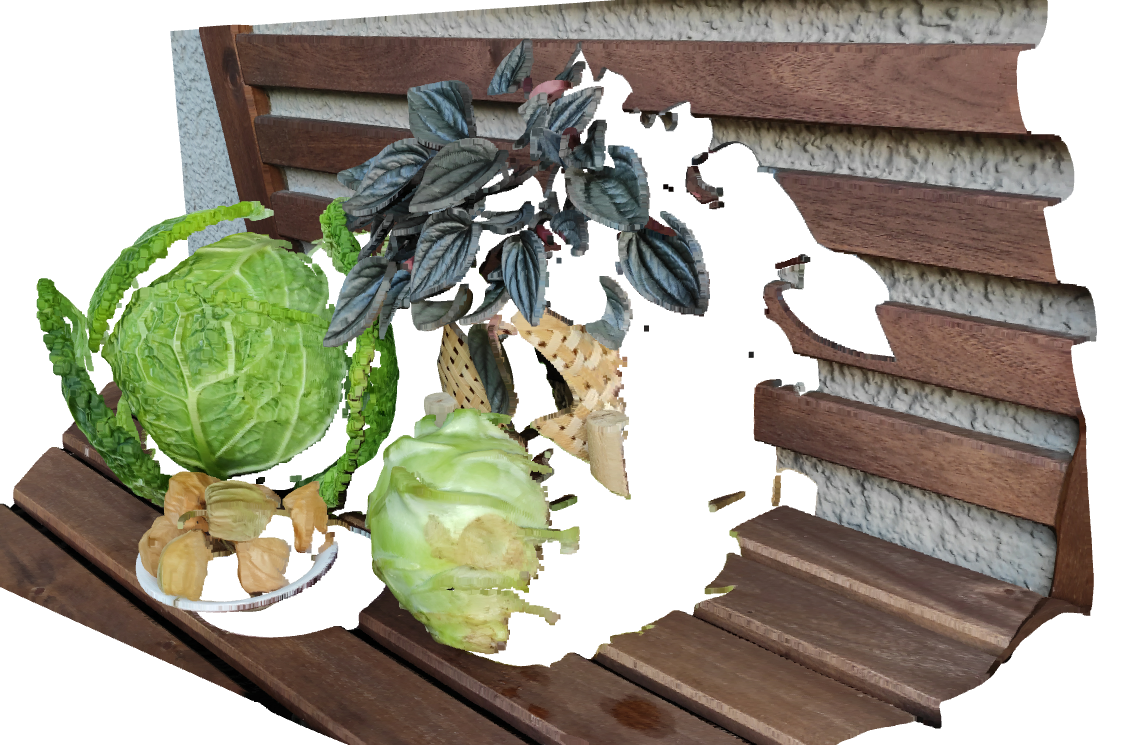} & 
            \includegraphics[width=0.2\textwidth]{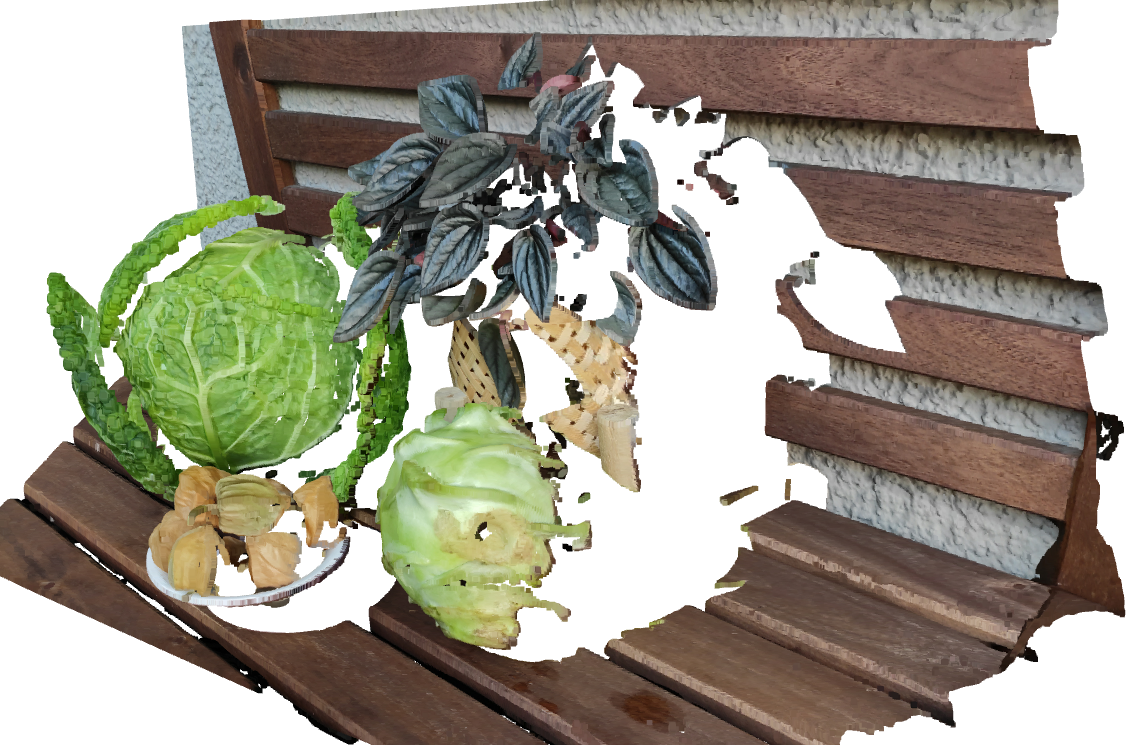} \\
            
            \includegraphics[width=0.2\textwidth]{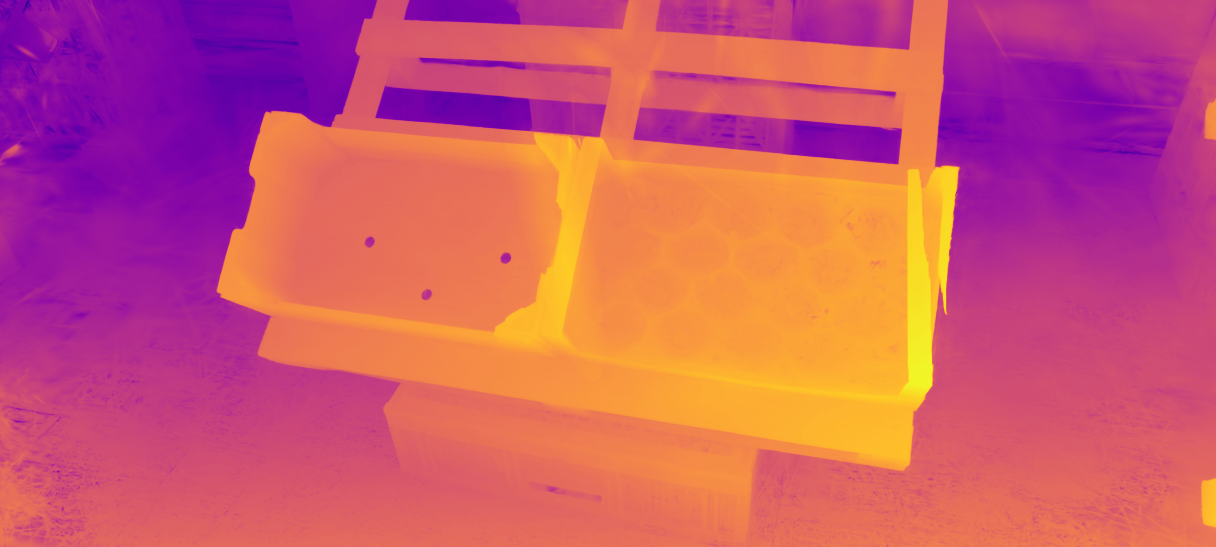} & 
            \includegraphics[width=0.2\textwidth]{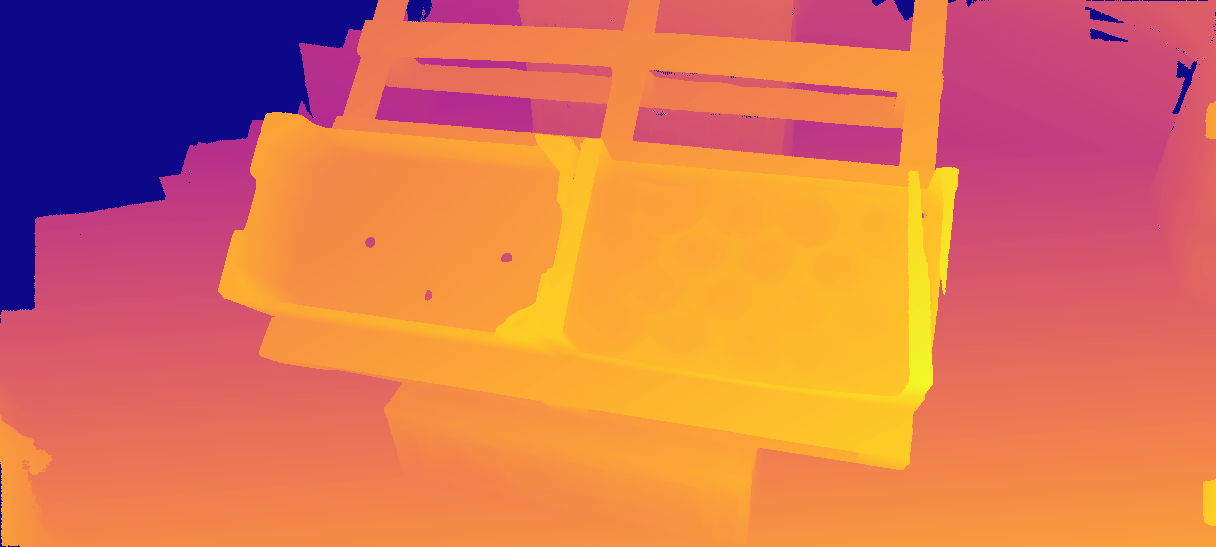} & 
            \includegraphics[width=0.2\textwidth]{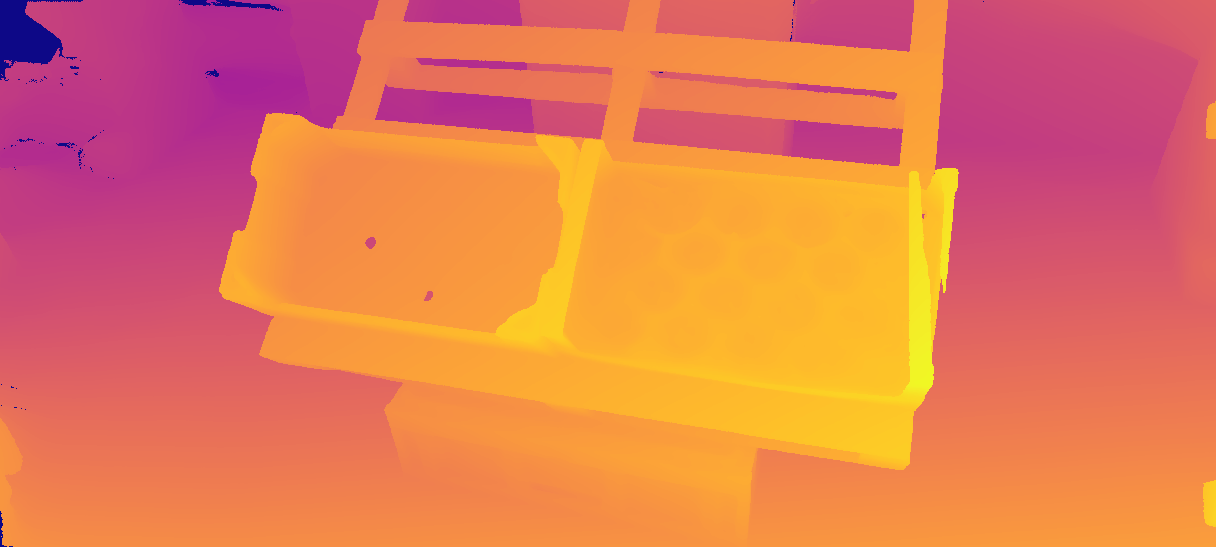} & 
            \includegraphics[width=0.2\textwidth]{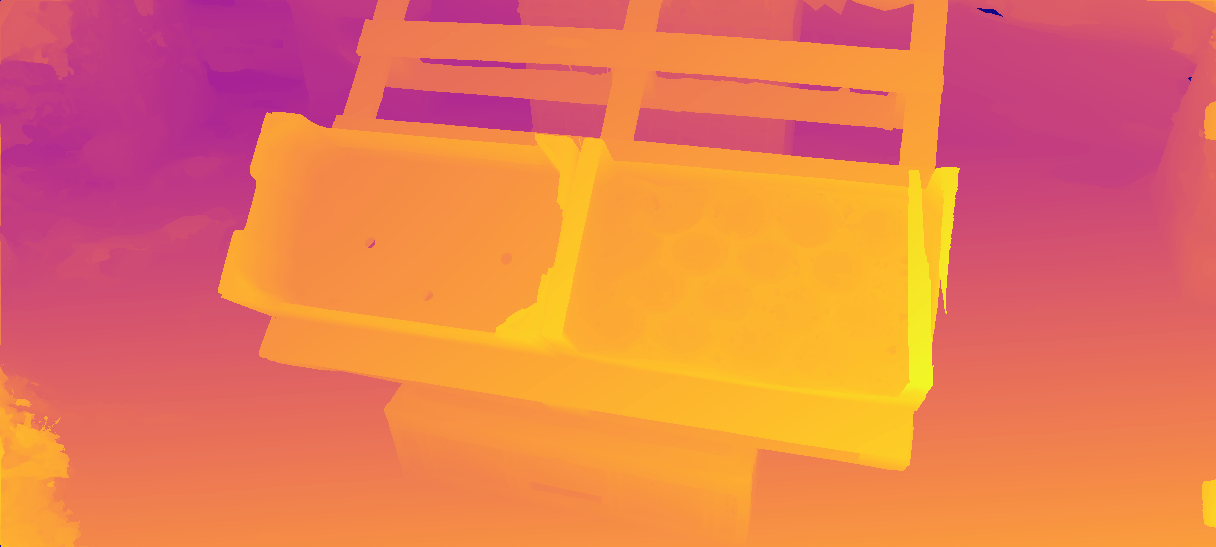} \\
                                        
            \includegraphics[width=0.2\textwidth]{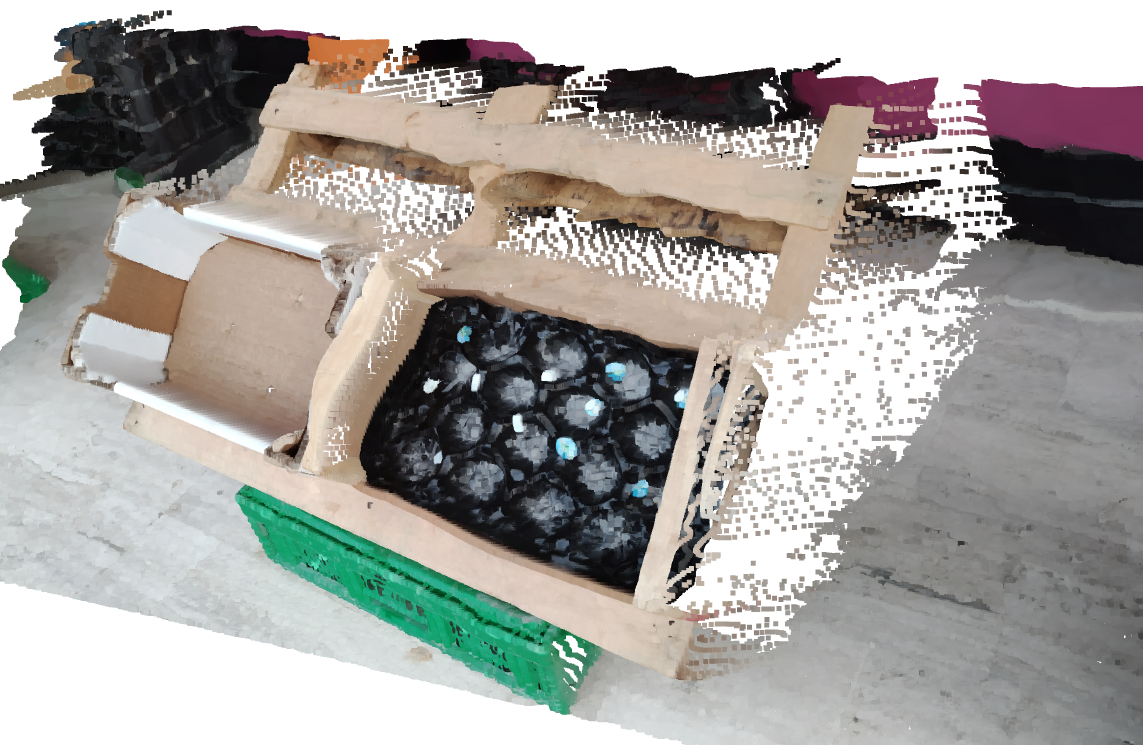} & 
            \includegraphics[width=0.2\textwidth]{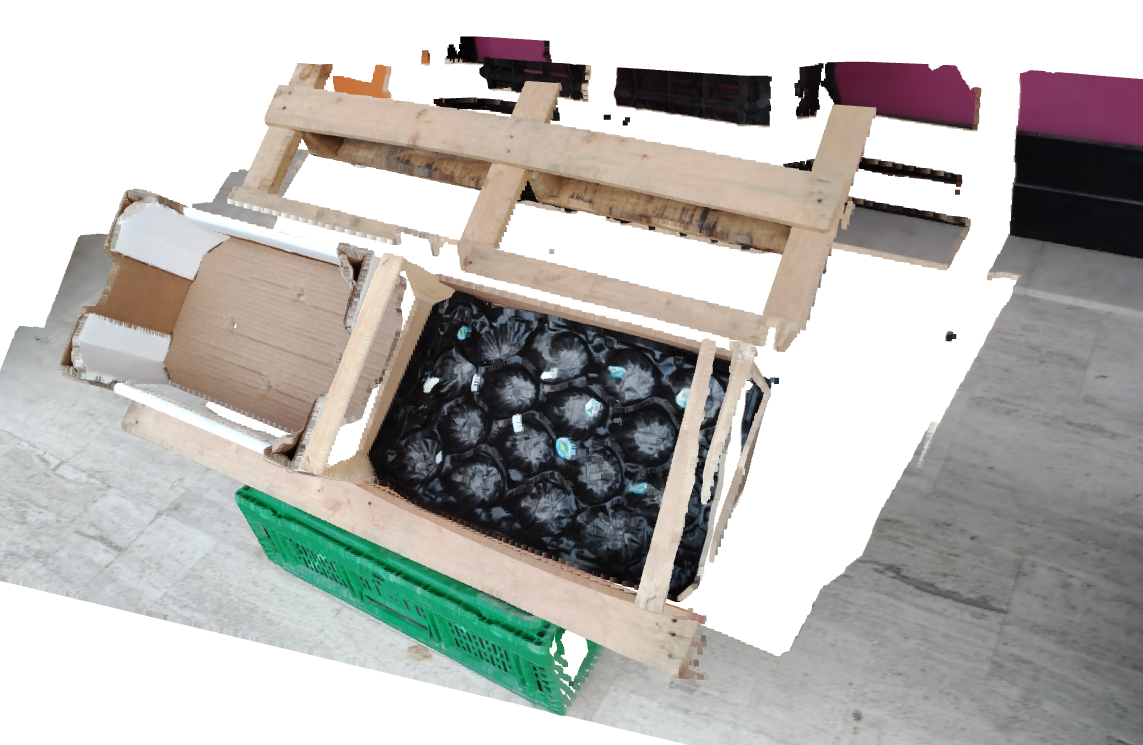} & 
            \includegraphics[width=0.2\textwidth]{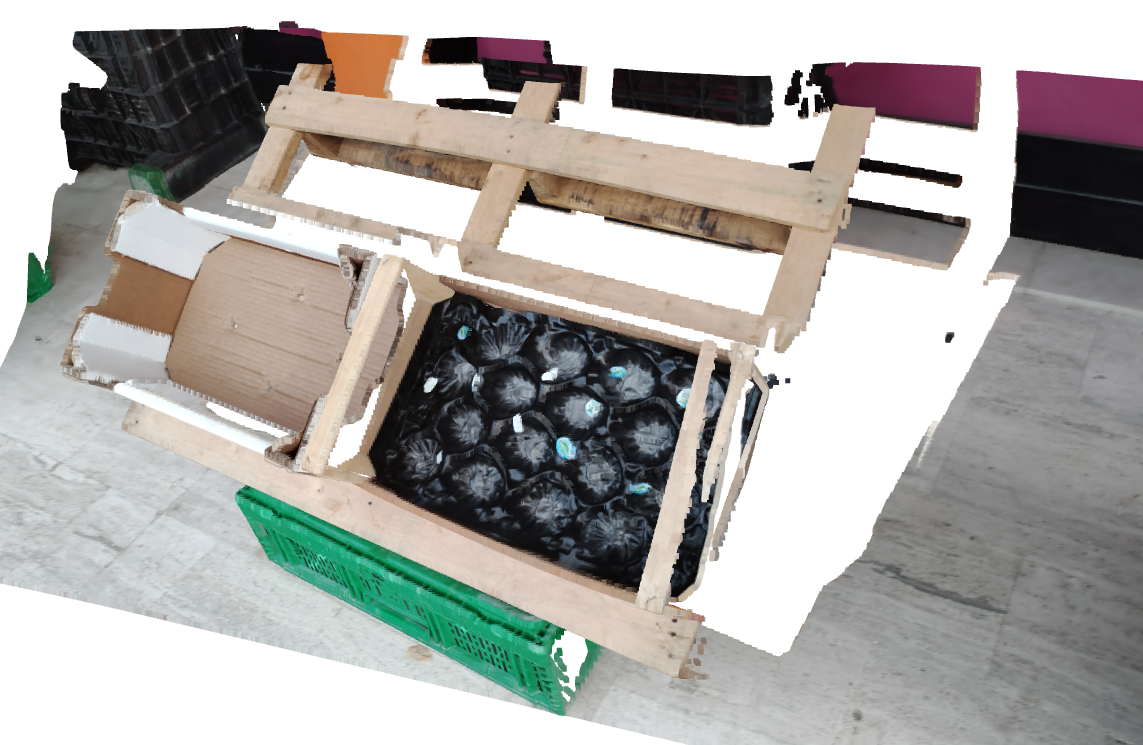} & 
            \includegraphics[width=0.2\textwidth]{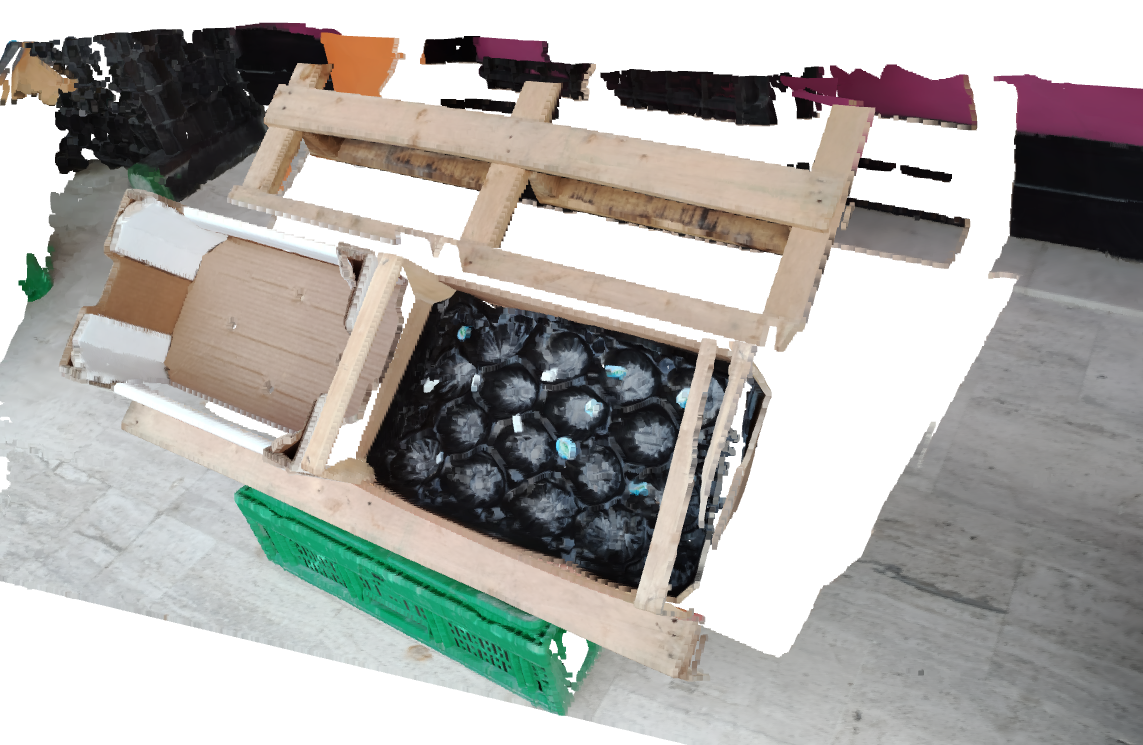} \\

    \end{tabular}
    \caption{\textbf{Reconstructed Geometry with Gaussian Splatting.} Qualitative examples of mesh reconstructions, renderings, and backprojected disparity images produced using the 3DGS, PGSR, 2DGS, and GOF methods. Note that, for illustrative clarity, spurious faces occluding the camera view in the GOF method have been manually removed.}
    \label{fig:mesh_rendering_comparison}
\end{figure*}

\begin{figure*}[!t]
\centering
\subfloat[High Observability]{\includegraphics[width=0.48\linewidth]{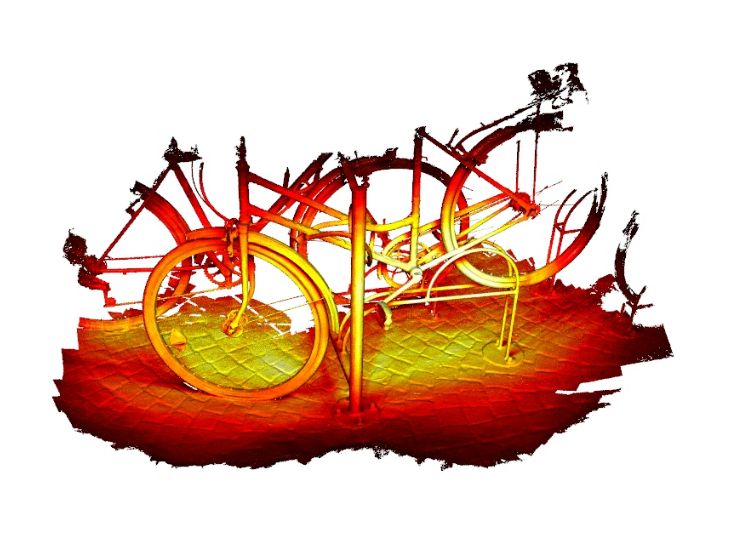}\label{fig:heatmap_1}}\hskip1ex
\subfloat[Low Observability]{\includegraphics[width=0.48\linewidth]{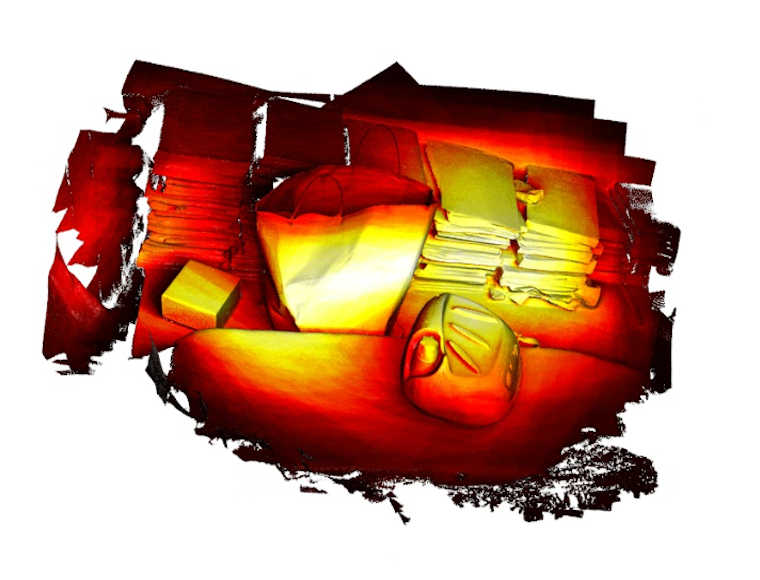}\label{fig:heatmap_2}}\hskip1ex
\caption{\textbf{Observability filtering:} Examples of the observability heatmaps where brighter colors represent vertices that have been seen by more cameras. Figure \ref{fig:heatmap_1} shows a high observability scene where an object is seen from many positions, and Figure \ref{fig:heatmap_2} shows a low-observability scene. Notice how there are many more dark areas in Figure \ref{fig:heatmap_2}, indicating they are poorly observed.}
\label{fig:heatmapExamples}
\end{figure*}
\section{Methodology}
\label{sec:methodology}
In this work, we investigate whether the recent advancements in 3D Gaussian Splatting can be used to synthesize novel views, reconstruct accurate surface geometries from images, and apply the results to the downstream task of stereo-matching with the RAFT-Stereo network \cite{lipson2021raft}.

\subsection{Stereo Dataset Rendering with 3DGS}\label{sec:3dgsStereo}
When rendering the disparity image from a 3DGS scene representation, the naive approach is to use the explicit representation of the gaussians and calculate the disparity based on the 3D location of the splat following a hard threshold on the opacity. Kerbl \etal \cite{Kerbl20233DGS} proposed a neural point-based approach to compute the color C of a pixel as:
\begin{equation} \label{eq:kerbl}
    C = \sum_{i \in N}c_i \alpha_i \prod_{j = 1}^{i-1}(1-\alpha_j)
\end{equation}
Where $c_i$ is the learned color of each point and $\alpha_i$ is the screen projected opacity of the splat. The naive depth rendering approach replaces the learned color of each point with the z-depth from the viewing transform. Since the new color represents a depth value, it can be converted to disparity $d$ in the rendered stereo pair with a virtual baseline $b$ and focal length $f$ as: 
\begin{equation} \label{eq:disparity}
    d = \frac{f b}{z_{depth}}
\end{equation}

This approach generates 'comet tails' smooth transitions at edges which should be sharp, similar to the rendered images used in the NeRF-Stereo work \cite{Tosi2023NeRFSupervisedDS}, showing little improvement over different Novel View Synthesis methods. Therefore, in order to improve the quality of the depth estimates for the stereo pairs, recent advances in mesh generation are considered \cite{Huang20242DGS, chen2024pgsr, Yu2024GaussianOF}.

The generated meshes can be efficiently rendered into a depth map and converted to disparity using Eq. \ref{eq:disparity}. As meshing-based methods are designed with accurate surface reconstruction in mind, artifacts such as comet tails should be reduced. To decide on the viewing transform for the dataset, we re-use the COLMAP \cite{schoenberger2016mvs}\cite{schoenberger2016sfm} poses along with the camera intrinsics. We use the same baselines as Tosi \etal \cite{Tosi2023NeRFSupervisedDS} to achieve the same disparity distribution in the dataset.

\subsection{Qualitative Assessment of Gaussian Splatting Methods}\label{sec:GSMeshQuality}

As mentioned, since the introduction of the 3D Gaussian Splatting method, there has been a rapid development of newer models with a focus of improving the underlying geometry. While it is possible to generate visually pleasing and convincing images with poor underlying geometry, training on noisy ground truth structural information will result in models of poor quality which produce blurred outputs. Therefore, we qualitatively evaluate three of the most recent and state-of-the-art Gaussian Splatting methods from which we can extract meshes explicitly. Namely we consider PGSR, 2DGS and GOF \cite{chen2024pgsr, Yu2024GaussianOF, Huang20242DGS}. Using each method, we reconstruct and render the dataset used in the NeRF-Stereo paper \cite{Tosi2023NeRFSupervisedDS}, and qualitatively investigate the reconstruction. For every method, we have used base settings for the rendering pipeline. Mesh rendering for PGSR and 2DGS is based on TSDF fusion and produces numerous floating artifacts, and a post-processing step is applied to keep only the largest cluster. However, while this improves the overall mesh quality, it also results in overfiltering, removing the finer correct details.

While each method can produce visually pleasing images, we find that they suffer from noisy and insufficient geometry reconstructions. In the qualitative comparison in Figure \ref{fig:mesh_rendering_comparison}, we observe that 2DGS provides a balanced tradeoff between finer details and geometric accuracy, whereas GOF contains significant artifacts and holes in the textureless areas. In contrast, PGSR results in the cleanest geometry with less artifacts, even though the back wall is excluded making the reconstruction very object centric. From Figure \ref{fig:mesh_rendering_comparison}, it is trivial to see the noisy results of plain 3DGS depth, with a significant amount of comet tails present. Based on these results, we decided to solely use the mesh reconstruction obtained from PGSR, as it produces the cleanest meshes. 

\subsection{Mesh Filtering}

Filtering the 3D structure directly can help with removing artifacts caused by the presence of insufficient information. Even though PGSR employs a filtering strategy masking all the surfaces where the camera rays intersect them at an angle greater than 80 degrees, the rasterized input depth maps contain artifacts due to under-reconstructed areas, causing incorrect surface estimation during the TSDF process. 


A prerequisite for accurate estimation of 3D structure from 2D views is the presence of parallax, meaning that each 3D point should be captured by cameras that are sufficiently displaced in relative pose. This principle is well aligned with our observation of the rendered meshes, where the quality of the 3D reconstruction was proportional to how well observed a given surface was given the camera system. Similarly, we have observed that the reconstructed visual quality was also proportional to how many training cameras have been observing a given surface during the novel view rendering. As such, we have turned these observations into a method which could be used to further filter existing meshes, but also provide valuable information about visual fidelity of each camera, which is especially critical considering that the proposed pipeline by Nerf-Stereo \cite{Tosi2023NeRFSupervisedDS} leverages the poses of training cameras. 

Given the rendered mesh and the known poses of all the training cameras relative to it, we calculate how many cameras have unoccluded view of every single vertex. A visual example can be seen in figure \ref{fig:heatmapExamples}. Once the meshes are enriched by the observability information, depth maps are rendered for every single training camera. Besides obtaining per-pixel depth, the observability is also propagated as additional attribute to every pixel. To rank the camera poses based on observability, we sum the observability across all the projected pixels. In summary, the observability based filtering can not only help in removing low-quality geometry from rendered meshes, but also ranks the training camera poses based on how well observed area they observe, providing an informed way to select fewer rendering poses, rather than choosing all as in \cite{Tosi2023NeRFSupervisedDS}.

\subsection{Transferring Expert Knowledge from FoundationStereo}

Motivated by our findings that the Novel View Synthesis methods produce poor geometry for in-the-wild datasets, as demonstrated in both the NeRF-Stereo paper and our explorations of the Gaussian Splatting meshing methods, we consider an orthogonal direction. Instead of relying on the computed geometry, we investigate the effectiveness of estimating depth using large-scale foundation models. Specifically, we use the FoundationStereo \cite{FoundationStereo} model to produce a pseudo disparity estimates for all rendered stereo pairs, which we use to train a RAFT-Stereo network, see Figure~\ref{fig:overview}. This can be seen as a way of performing knowledge distillation, as we are effectively training a smaller-scale network on predictions of a large-scale network, which would be prohibitively slow in real-time applications. It is also worth noting that FoundationStereo was trained purely on synthetic data, eliminating any concerns of data leakage from the test datasets. This approach is similar, but distinct, from the prior GS2Mesh work \cite{GS2MESH} which used depth estimates to produce detailed geometry of scenes rendered with Gaussian Splatting.

\begin{figure*}[t]
  \centering
  \includegraphics[width=\textwidth]{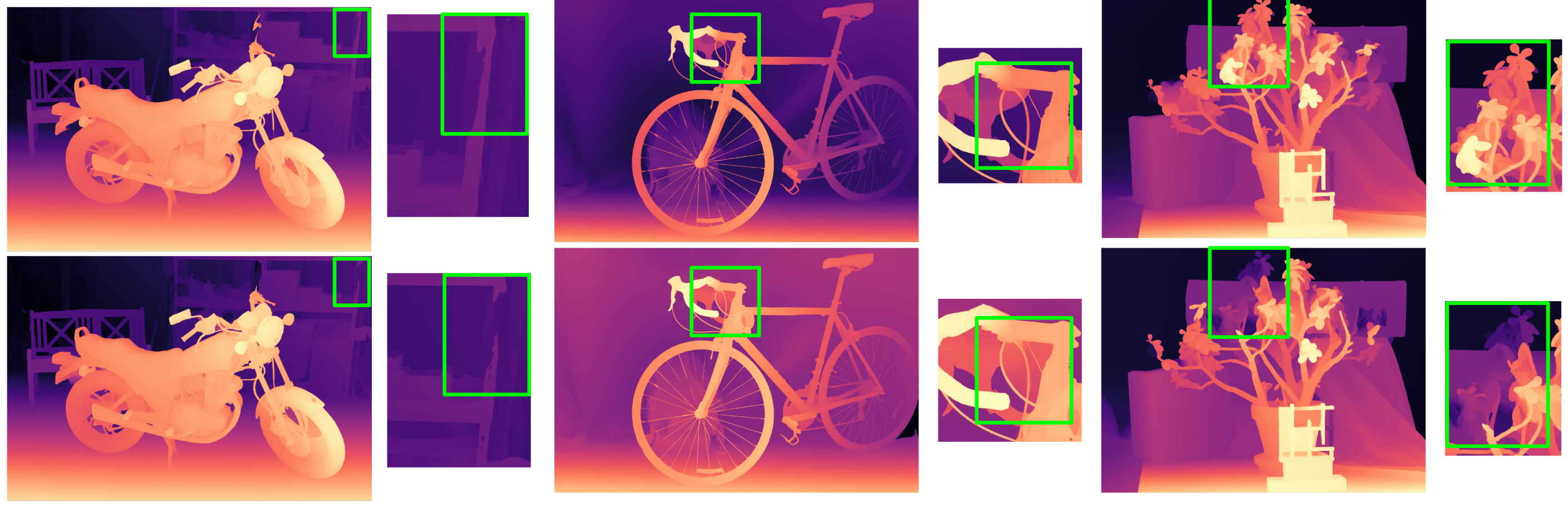}
  \caption{\textbf{Qualitative Comparison of RAFT-Stereo and 3DGS + FS.} We make a visual comparison between RAFT-Stereo SceneFlow checkpoint, and our finetuned version on dataset generated using our proposed pipeline leveraging 3DGS for stereo images and FoundationStereo for ground truth pseudolabels. Our proposed 3DGS + FS method is shown in the top row, and predictions from the baseline RAFT-Stereo method are shown in the bottom row. Improvements can be observed in thin structures and complex depth discontinuities}
  \label{fig:qualitative_comparison}
\end{figure*}

%% file: sec/4_experiments.tex
\section{Experimental Design}
Disparity estimation models pretrained on synthetic datasets possess impressive reasoning skills about spatial geometry, but still fall short due to the sim2real gap in the image space. However, only a small amount of real data is required to improve the matching accuracy on real data, as demonstrated by RAFT-Stereo. We follow the exact finetuning procedure and augmentations outlined in their work, designed for finetuning SceneFlow checkpoint on smaller scale, but high quality real data. We further consider this training setup appropriate, as the scenes from Nerf-Stereo dataset looks visually similar to Middlebury style datasets. While RAFT-Stereo was fine-tuned on a Middlebury 2014 dataset \cite{Scharstein2014HighResolutionSD}, we generate the training dataset following the rendering procedure of Nerf-Stereo \cite{Tosi2023NeRFSupervisedDS} outlined in Section~\ref{sec:3dgsStereo}. 


As the extracted poses from the camera systems using COLMAP are used as anchors to generate the stereo pairs, their observations vary in quality. Our experiments show that using the observability score per camera is a great predictor of stereo image quality, ensuring that the very left camera image observes the best reconstructed area of the scene. 
Given the five cameras with maximized observability, we generate three baselines following Nerf-Stereo, ensuring disparity distribution is similar to the test and validation datasets. We also show results when training using the 3DGS rendered images and FoundationStereo pseudo depth estimates. For all the experiments, regardless of the method, we select the same scenes and camera locations to ensure a fair comparison.

 Lastly, we compare the performance to the RAFT-Stereo model trained on the Sceneflow dataset as well as NeRF-Stereo. It is worth highlighting that NeRF-Stereo was trained from scratch with a trinocular and photometric loss, depending on filtering pixels with poor geometry estimates based on their transparency\footnote{Incorrectly called Ambient Occlusion (AO) filtering in the NeRF-Stereo paper, and in this work referred to as $\alpha$ filtering.}. We also report NeRF-Stereo results obtained from trying to reproduce the training process using an unofficial code reproduction\footnote{\url{https://github.com/husheng12345/Unofficial-NeRF-Supervised-Deep-Stereo}} by the authors of the RAFT-Stereo paper, as the original stereo-network training code is not available. 

We follow the NeRF-Stereo evaluation protocol and report results on 200 stereo images from KITTI 2015 \cite{KITTI2015}, 15 stereo pairs from Middlebury v3 training set (Midd-T) \cite{Scharstein2014HighResolutionSD} and 27 pairs from ETH3D \cite{Schps2017AMS}. The best checkpoint per method is chosen using a suite of validation datasets, specifically 194 stereo images from KITTI 2012 \cite{Geiger2012AreWR}, 13 additional images from the training set of Middlebury v3 (Midd-A) at Full, Half, and Quarter resolutions (F, H, Q), and the Middlebury 2021 (Midd-21) dataset \cite{Scharstein2014HighResolutionSD}. Performance is measured by determining the ratio of pixels with a disparity error larger than $\tau$ pixels. $\tau$ is dataset dependent and set to 1 for ETH3D, 2 for Middlebury, and 3 for KITTI, following common practice in the stereo matching field.


\begin{table*}[]
\centering
\footnotesize 
\renewcommand{\arraystretch}{1.5} 
\begin{tabular}{l c c | c c | c c c c c c | c c }

 & & & \multicolumn{2}{c|}{\textbf{KITTI-15}} & \multicolumn{6}{c|}{\textbf{Midd-T}} & \multicolumn{2}{c}{\textbf{ETH3D}} \\ 

\textbf{ } & & & \multicolumn{2}{c|}{($>$3px)} & \multicolumn{2}{c}{F ($>$2px)}& \multicolumn{2}{c}{H ($>$2px)}& \multicolumn{2}{c|}{Q ($>$2px)} &  \multicolumn{2}{c}{($>$1px)} \\ 

 \textbf{Method} & \textbf{Filtering} & \textbf{Fine-Tuned} & \textbf{All} & \textbf{Noc} & \textbf{All} & \textbf{Noc} & \textbf{All} & \textbf{Noc} & \textbf{All} & \textbf{Noc} & \textbf{All} & \textbf{Noc} \\ \hline
RAFT-Stereo & - & \xmark & \fcolorbox{silver}{silver!10}{\underline{5.46}} & \fcolorbox{silver}{silver!10}{\underline{5.27}} & \fcolorbox{silver}{silver!10}{\underline{15.72}} & 12.00 & 11.23 & 8.67 & 10.52 & 7.42 & 2.61 & 2.29 \\ \hline
NeRF-Stereo & $\alpha$ & \xmark & \fcolorbox{gold}{gold!10}{\textbf{5.41}} & \fcolorbox{gold}{gold!10}{\textbf{5.23}} & 16.45 & 12.08 & \fcolorbox{silver}{silver!10}{\underline{9.67}} & \fcolorbox{gold}{gold!10}{\textbf{6.42}} & \fcolorbox{gold}{gold!10}{\textbf{8.05}} & \fcolorbox{gold}{gold!10}{\textbf{4.82}} & 2.94 & 2.23 \\
NeRF-Stereo* & $\alpha$ &\xmark & 6.65 & 6.22 & 20.10 & 15.43 & 12.82 & 8.89 & 11.18 & 7.68 & 4.37 & 3.91 \\  
NeRF-Stereo & $\alpha$ &\cmark & 5.63 & 5.41 & 15.89 & \fcolorbox{silver}{silver!10}{\underline{11.72}}  & 11.03 & 7.84 & 9.71 & 6.54 & 3.48 & 3.00 \\  \hline
3DGS & - & \cmark  & 5.77 & 5.59 & 22.74 & 19.97 & 15.53 & 11.96 & 10.41 & 7.11 & 4.65 & 4.28 \\

PGSR & - & \cmark & 6.26 & 6.09 & 16.64 & 12.81 & 11.23 & 8.26 & 10.32 & 7.52 & 3.19 & 3.04 \\
PGSR & obs. & \cmark & 6.51 & 6.34 & 17.14 & 13.34 & 11.75 & 8.69 & 9.83 & 7.26 & 3.08 & 2.89 \\ \hline

3DGS + FS & - &  \cmark & 5.52 & 5.31 & \fcolorbox{gold}{gold!10}{\textbf{14.78}} & \fcolorbox{gold}{gold!10}{\textbf{11.25}} & \fcolorbox{gold}{gold!10}{\textbf{9.62}} & \fcolorbox{silver}{silver!10}{\underline{6.80}} & \fcolorbox{silver}{silver!10}{\underline{9.00}} & \fcolorbox{silver}{silver!10}{\underline{6.18}} & \fcolorbox{gold}{gold!10}{\textbf{2.35}} & \fcolorbox{gold}{gold!10}{\textbf{2.14}} \\
\end{tabular}
\normalsize 
\caption{\textbf{Zero-Shot Generalization Benchmark.} We compare the zero-shot performance of various synthetic dataset generation methods. All results are obtained from the RAFT-Stereo model. We report both the original NeRF-Stereo results as well as the reproduced results (denoted by *), with the model in both cases trained from scratch. The Fine-tuned column indicates whether the model was initialized with the SceneFlow checkpoint. The filtering column indicates which, if any, filtering methods are used on the data. Obs.~indicates observability filtering was used to select scenes, whereas $\alpha$ indicates the pixel transparency was thresholded (denoted AO filtering in the NeRF-Stereo paper). 3DGS + FS denotes 3DGS with FoundationStereo \cite{FoundationStereo} depth estimates. We highlight the \fcolorbox{gold}{gold!10}{\textbf{best}} and \fcolorbox{silver}{silver!10}{\underline{second-best}} results. \label{tab:results}} \end{table*}

\section{Experimental Results}
We report our results in Table~\ref{tab:results}. We find that finetuning on the 3DGS and PGSR based datasets is comparable to the performance of the reproduced NeRF-Stereo training paradigm from random initialization on their full dataset. However, none of them (including the reproduced NeRF-Stereo) are comparable to the RAFT-Stereo baseline, which was solely trained on SceneFlow \cite{Mayer2015ALD}. This is a clear indication that the reconstructed geometry in the Gaussian Splatting-based methods is of too poor quality, as discussed in Section~\ref{sec:GSMeshQuality}. We also find that the observation filtering used with PGSR does not consistently improve performance. Instead, performance is only improved on Midd-T at quarter resolution as well as on the ETH3D dataset. 

Interestingly, we observe that when fine-tuning using the NeRF-Stereo renderings, the performance becomes significantly better and even beats the RAFT-Stereo baseline on the Midd-T dataset. We attribute this to the $\alpha$-filtering used by NeRF-Stereo, which aggressively filters out pixels with poor underlying geometry. Lastly, we find that our expert knowledge transfer setup (3DGS + FS) outperforms all other methods on the Midd-T and ETH3D datasets, and is close to matching the RAFT-Stereo performance on the KITTI-15 dataset. We visually examine the predicted disparity maps from the baseline RAFT-Stereo and the proposed 3DGS+FS, see Figure~\ref{fig:qualitative_comparison}. We find that our proposed methods are better at recovering very thin structures (such as the brake cables on the bicycle) as well as complex layered scenes (such as the flowers on the table). It is worth noting that the original reported results from NeRF-Stereo do outperform 3DGS+FS on KITTI-15 and parts of Midd-T. However, as the results are not reproducible, we argue that they are not fully representative.

%% file: sec/5_discussion_conclusion.tex
\section{Dataset and Training Instabilities}

Throughout the proposed experiments, we encountered several instances of instability when constructing the synthetic datasets and training the stereo networks.

Firstly, we encountered several problems with the NeRF-Stereo dataset. As the data is collected with handheld cameras being moved around, there is a considerable amount of motion blur and poor quality images, which we found the Gaussian Splatting methods are quite sensitive towards. We also found that the provided COLMAP poses result in poor reconstruction with Gaussian Splatting-based methods, as the principal point was refined and did not lie in the center of the image. This is a known problem\footnote{\url{https://github.com/graphdeco-inria/gaussian-splatting/issues/144\#issuecomment-1786762565}} and was rectified by recomputing COLMAP without refining the principal points. While it is practically possible to convert images with non-centred principal points by simple cropping and adjusting the camera paramters, we have chosen to rerun the scenes with the provided \textit{convert.py} script common in gaussian splatting repositories, as that would align with the core idea of generating custom training datasets only from a set of freely captured images.  However, this in turn resulted in COLMAP failing to estimate poses for several scenes, even when attempted multiple times. 

Secondly, we found that the scene reconstruction with 3D Gaussian Splatting based methods can be prohibitively expensive, with some scenes failing to be reconstructed due to insufficient VRAM even when using an A100 with 40GB VRAM. On average, 10\% of scenes or meshes were not reconstructed with 3DGS, GOF, 2DGS, or PGSR.


\section{Discussion}
Despite the excellent performance of the latest 3DGS-based methods on popular 3D reconstruction benchmarks from multi-view images, our study has revealed that they are still insufficient for in-the-wild scenarios. Although it was impossible to evaluate the accuracy of the 3D reconstructions due to the absence of associated ground truth data, assessing accuracy by using a proxy downstream task can be equally or even more valuable for judging the robustness of such methods. As such, the poor performance when training on the data rendered from meshes has been unexpected and suggests that the current benchmark performance is not representative of in-the-wild settings, which can be regarded as a valuable result and inspire future works. On the other hand, the FoundationStereo performance has shown to be representative even in a zero-shot setup, suggesting that the methodology from GS2Mesh \cite{GS2MESH} can be further improved. Moreover, while it could be argued that 3DGS-based methods could be replaced with NeRFs, it should be considered that 3DGS-based methods offer superior rendering speeds at comparable visual fidelity, which is essential when rendering large-scale datasets.

\section{Conclusion}
In this paper, we have investigated the feasibility of using 3D Gaussian Splatting-based methods to generate synthetic datasets to train a stereo network. Through qualitative assessment of state-of-the-art meshing-based methods, we find that the reconstructed geometry is consistently of insufficient quality with artifacts, noise, and holes present. This is consequently reflected in poor performance when fine-tuning a RAFT-Stereo network, even when applying filtering to only use the most well-observed parts of the meshes. We further show that by utilizing an expert knowledge transfer setup, where depth estimates from the FoundationStereo model are used as pseudo ground truth, we achieve much better performance. We attribute the success of this methodology to the fact that Gaussian Splatting methods capture high-fidelity visual images even if they are unable to capture underlying geometry due to the limitations of traditional algorithms. On the other hand, superior zero-shot performance of FoundationStereo can infer accurate geometry even from very limited visual signal. We consider our results to be a promising avenue for future work on generating synthetic stereo datasets and transferring expert knowledge into small and lightweight stereo networks, while offering unprecedented flexibility of capturing custom datasets from just freely captured images.

\textbf{Funding} This research was funded by Innovation Fund Denmark, grant number 3129-00060B.

%% file: main.bbl
\begin{thebibliography}{36}
\providecommand{\natexlab}[1]{#1}
\providecommand{\url}[1]{\texttt{#1}}
\expandafter\ifx\csname urlstyle\endcsname\relax
  \providecommand{\doi}[1]{doi: #1}\else
  \providecommand{\doi}{doi: \begingroup \urlstyle{rm}\Url}\fi

\bibitem[Barron et~al.(2021)Barron, Mildenhall, Tancik, Hedman, Martin-Brualla, and Srinivasan]{Barron2021MipNeRFAM}
Jonathan~T. Barron, Ben Mildenhall, Matthew Tancik, Peter Hedman, Ricardo Martin-Brualla, and Pratul~P. Srinivasan.
\newblock Mip-nerf: A multiscale representation for anti-aliasing neural radiance fields.
\newblock \emph{2021 IEEE/CVF International Conference on Computer Vision (ICCV)}, pages 5835--5844, 2021.

\bibitem[Barron et~al.(2023)Barron, Mildenhall, Verbin, Srinivasan, and Hedman]{Barron2023ZipNeRFAG}
Jonathan~T. Barron, Ben Mildenhall, Dor Verbin, Pratul~P. Srinivasan, and Peter Hedman.
\newblock Zip-nerf: Anti-aliased grid-based neural radiance fields.
\newblock \emph{2023 IEEE/CVF International Conference on Computer Vision (ICCV)}, pages 19640--19648, 2023.

\bibitem[Chang and Chen(2018)]{chang2018pyramid}
Jia-Ren Chang and Yong-Sheng Chen.
\newblock Pyramid stereo matching network.
\newblock In \emph{Proceedings of the IEEE conference on computer vision and pattern recognition}, pages 5410--5418, 2018.

\bibitem[Chen et~al.(2024)Chen, Li, Ye, Wang, Xie, Zhai, Wang, Liu, Bao, and Zhang]{chen2024pgsr}
Danpeng Chen, Hai Li, Weicai Ye, Yifan Wang, Weijian Xie, Shangjin Zhai, Nan Wang, Haomin Liu, Hujun Bao, and Guofeng Zhang.
\newblock Pgsr: Planar-based gaussian splatting for efficient and high-fidelity surface reconstruction.
\newblock 2024.

\bibitem[Dai et~al.(2024)Dai, Xu, Xie, Liu, Wang, and Xu]{Dai2024HighqualitySR}
Pinxuan Dai, Jiamin Xu, Wenxiang Xie, Xinguo Liu, Huamin Wang, and Weiwei Xu.
\newblock High-quality surface reconstruction using gaussian surfels.
\newblock In \emph{International Conference on Computer Graphics and Interactive Techniques}, 2024.

\bibitem[Eftekhar et~al.(2021)Eftekhar, Sax, Bachmann, Malik, and Zamir]{Eftekhar2021OmnidataAS}
Ainaz Eftekhar, Alexander Sax, Roman Bachmann, Jitendra Malik, and Amir Zamir.
\newblock Omnidata: A scalable pipeline for making multi-task mid-level vision datasets from 3d scans.
\newblock \emph{2021 IEEE/CVF International Conference on Computer Vision (ICCV)}, pages 10766--10776, 2021.

\bibitem[Gehrig et~al.(2021)Gehrig, Aarents, Gehrig, and Scaramuzza]{Gehrig2021DSECAS}
Mathias Gehrig, Willem Aarents, Daniel Gehrig, and Davide Scaramuzza.
\newblock Dsec: A stereo event camera dataset for driving scenarios.
\newblock \emph{IEEE Robotics and Automation Letters}, 6:\penalty0 4947--4954, 2021.

\bibitem[Geiger et~al.(2012)Geiger, Lenz, and Urtasun]{Geiger2012AreWR}
Andreas Geiger, Philip Lenz, and Raquel Urtasun.
\newblock Are we ready for autonomous driving? the kitti vision benchmark suite.
\newblock \emph{2012 IEEE Conference on Computer Vision and Pattern Recognition}, pages 3354--3361, 2012.

\bibitem[Geiger et~al.(2013)Geiger, Lenz, Stiller, and Urtasun]{Geiger2013VisionMR}
Andreas Geiger, Philip Lenz, Christoph Stiller, and Raquel Urtasun.
\newblock Vision meets robotics: The kitti dataset.
\newblock \emph{The International Journal of Robotics Research}, 32:\penalty0 1231 -- 1237, 2013.

\bibitem[Gjerde et~al.(2024)Gjerde, Slez{\'a}k, Haurum, and Moeslund]{FilipCVPRW}
Magnus~Kaufmann Gjerde, Filip Slez{\'a}k, Joakim~Bruslund Haurum, and Thomas~B. Moeslund.
\newblock From ne{RF} to 3{DGS}: A leap in stereo dataset quality?
\newblock In \emph{Synthetic Data for Computer Vision Workshop @ CVPR 2024}, 2024.

\bibitem[Hirschmuller(2005)]{hirschmuller2005accurate}
Heiko Hirschmuller.
\newblock Accurate and efficient stereo processing by semi-global matching and mutual information.
\newblock In \emph{2005 IEEE computer society conference on computer vision and pattern recognition (CVPR'05)}, pages 807--814. IEEE, 2005.

\bibitem[Hu et~al.(2024)Hu, Yin, Zhang, Cai, Long, Chen, Wang, Yu, Shen, and Shen]{hu2024metric3d}
Mu Hu, Wei Yin, Chi Zhang, Zhipeng Cai, Xiaoxiao Long, Hao Chen, Kaixuan Wang, Gang Yu, Chunhua Shen, and Shaojie Shen.
\newblock Metric3d v2: A versatile monocular geometric foundation model for zero-shot metric depth and surface normal estimation.
\newblock \emph{IEEE Transactions on Pattern Analysis and Machine Intelligence}, 2024.

\bibitem[Huang et~al.(2024)Huang, Yu, Chen, Geiger, and Gao]{Huang20242DGS}
Binbin Huang, Zehao Yu, Anpei Chen, Andreas Geiger, and Shenghua Gao.
\newblock 2d gaussian splatting for geometrically accurate radiance fields.
\newblock \emph{ArXiv}, abs/2403.17888, 2024.

\bibitem[Huang et~al.(2021)Huang, Gu, Li, and Yu]{huang2021stereo}
Zedong Huang, Jinan Gu, Jing Li, and Xuefei Yu.
\newblock A stereo matching algorithm based on the improved psmnet.
\newblock \emph{Plos one}, 16\penalty0 (8):\penalty0 e0251657, 2021.

\bibitem[Kerbl et~al.(2023)Kerbl, Kopanas, Leimkuehler, and Drettakis]{Kerbl20233DGS}
Bernhard Kerbl, Georgios Kopanas, Thomas Leimkuehler, and George Drettakis.
\newblock 3d gaussian splatting for real-time radiance field rendering.
\newblock \emph{ACM Transactions on Graphics (TOG)}, 42:\penalty0 1 -- 14, 2023.

\bibitem[Li et~al.(2021)Li, Liu, Drenkow, Ding, Creighton, Taylor, and Unberath]{li2021revisiting}
Zhaoshuo Li, Xingtong Liu, Nathan Drenkow, Andy Ding, Francis~X Creighton, Russell~H Taylor, and Mathias Unberath.
\newblock Revisiting stereo depth estimation from a sequence-to-sequence perspective with transformers.
\newblock In \emph{Proceedings of the IEEE/CVF international conference on computer vision}, pages 6197--6206, 2021.

\bibitem[Ling et~al.(2024)Ling, Sun, Sun, Xu, and Li]{ADFactory}
Han Ling, Quansen Sun, Yinghui Sun, Xian Xu, and Xingfeng Li.
\newblock Adfactory: An effective framework for generalizing optical flow with nerf.
\newblock In \emph{2024 IEEE/CVF Conference on Computer Vision and Pattern Recognition (CVPR)}, pages 20591--20600, 2024.

\bibitem[Lipson et~al.(2021)Lipson, Teed, and Deng]{lipson2021raft}
Lahav Lipson, Zachary Teed, and Jia Deng.
\newblock Raft-stereo: Multilevel recurrent field transforms for stereo matching.
\newblock In \emph{2021 International Conference on 3D Vision (3DV)}, pages 218--227. IEEE, 2021.

\bibitem[Mayer et~al.(2015)Mayer, Ilg, H{\"a}usser, Fischer, Cremers, Dosovitskiy, and Brox]{Mayer2015ALD}
Nikolaus Mayer, Eddy Ilg, Philip H{\"a}usser, Philipp Fischer, Daniel Cremers, Alexey Dosovitskiy, and Thomas Brox.
\newblock A large dataset to train convolutional networks for disparity, optical flow, and scene flow estimation.
\newblock \emph{2016 IEEE Conference on Computer Vision and Pattern Recognition (CVPR)}, pages 4040--4048, 2015.

\bibitem[Menze et~al.(2015)Menze, Heipke, and Geiger]{KITTI2015}
M. Menze, C. Heipke, and A. Geiger.
\newblock Joint 3d estimation of vehicles and scene flow.
\newblock \emph{ISPRS Annals of the Photogrammetry, Remote Sensing and Spatial Information Sciences}, II-3/W5:\penalty0 427--434, 2015.

\bibitem[Mildenhall et~al.(2020)Mildenhall, Srinivasan, Tancik, Barron, Ramamoorthi, and Ng]{Mildenhall2020NeRF}
Ben Mildenhall, Pratul~P. Srinivasan, Matthew Tancik, Jonathan~T. Barron, Ravi Ramamoorthi, and Ren Ng.
\newblock Nerf.
\newblock \emph{Communications of the ACM}, 65:\penalty0 99 -- 106, 2020.

\bibitem[Oquab et~al.(2023)Oquab, Darcet, Moutakanni, Vo, Szafraniec, Khalidov, Fernandez, Haziza, Massa, El-Nouby, et~al.]{oquab2023dinov2}
Maxime Oquab, Timoth{\'e}e Darcet, Th{\'e}o Moutakanni, Huy Vo, Marc Szafraniec, Vasil Khalidov, Pierre Fernandez, Daniel Haziza, Francisco Massa, Alaaeldin El-Nouby, et~al.
\newblock Dinov2: Learning robust visual features without supervision.
\newblock \emph{arXiv preprint arXiv:2304.07193}, 2023.

\bibitem[Safadoust et~al.(2024)Safadoust, Tosi, G{\"u}ney, and Poggi]{SelfEvlovling3DGS}
Sadra Safadoust, Fabio Tosi, Fatma G{\"u}ney, and Matteo Poggi.
\newblock Self-evolving depth-supervised 3d gaussian splatting from rendered stereo pairs.
\newblock In \emph{British Machine Vision Conference (BMVC)}, 2024.

\bibitem[Scharstein et~al.(2014)Scharstein, Hirschm{\"u}ller, Kitajima, Krathwohl, Nesic, Wang, and Westling]{Scharstein2014HighResolutionSD}
Daniel Scharstein, Heiko Hirschm{\"u}ller, York Kitajima, Greg Krathwohl, Nera Nesic, Xi Wang, and Porter Westling.
\newblock High-resolution stereo datasets with subpixel-accurate ground truth.
\newblock In \emph{German Conference on Pattern Recognition}, 2014.

\bibitem[Sch\"{o}nberger and Frahm(2016)]{schoenberger2016sfm}
Johannes~Lutz Sch\"{o}nberger and Jan-Michael Frahm.
\newblock Structure-from-motion revisited.
\newblock In \emph{Conference on Computer Vision and Pattern Recognition (CVPR)}, 2016.

\bibitem[Sch\"{o}nberger et~al.(2016)Sch\"{o}nberger, Zheng, Pollefeys, and Frahm]{schoenberger2016mvs}
Johannes~Lutz Sch\"{o}nberger, Enliang Zheng, Marc Pollefeys, and Jan-Michael Frahm.
\newblock Pixelwise view selection for unstructured multi-view stereo.
\newblock In \emph{European Conference on Computer Vision (ECCV)}, 2016.

\bibitem[Sch{\"o}ps et~al.(2017)Sch{\"o}ps, Sch{\"o}nberger, Galliani, Sattler, Schindler, Pollefeys, and Geiger]{Schps2017AMS}
Thomas Sch{\"o}ps, Johannes~L. Sch{\"o}nberger, S. Galliani, Torsten Sattler, Konrad Schindler, Marc Pollefeys, and Andreas Geiger.
\newblock A multi-view stereo benchmark with high-resolution images and multi-camera videos.
\newblock \emph{2017 IEEE Conference on Computer Vision and Pattern Recognition (CVPR)}, pages 2538--2547, 2017.

\bibitem[Teed and Deng(2021)]{teed2021raft}
Zachary Teed and Jia Deng.
\newblock Raft-3d: Scene flow using rigid-motion embeddings.
\newblock In \emph{Proceedings of the IEEE/CVF conference on computer vision and pattern recognition}, pages 8375--8384, 2021.

\bibitem[Tosi et~al.(2023)Tosi, Tonioni, Gregorio, and Poggi]{Tosi2023NeRFSupervisedDS}
Fabio Tosi, Alessio Tonioni, Daniele~De Gregorio, and Matteo Poggi.
\newblock Nerf-supervised deep stereo.
\newblock \emph{2023 IEEE/CVF Conference on Computer Vision and Pattern Recognition (CVPR)}, pages 855--866, 2023.

\bibitem[Wang et~al.(2024)Wang, Xu, Jia, and Yang]{wang2024selective}
Xianqi Wang, Gangwei Xu, Hao Jia, and Xin Yang.
\newblock Selective-stereo: Adaptive frequency information selection for stereo matching.
\newblock In \emph{Proceedings of the IEEE/CVF Conference on Computer Vision and Pattern Recognition}, pages 19701--19710, 2024.

\bibitem[Weinzaepfel et~al.(2023)Weinzaepfel, Lucas, Leroy, Cabon, Arora, Br{\'e}gier, Csurka, Antsfeld, Chidlovskii, and Revaud]{weinzaepfel2023croco}
Philippe Weinzaepfel, Thomas Lucas, Vincent Leroy, Yohann Cabon, Vaibhav Arora, Romain Br{\'e}gier, Gabriela Csurka, Leonid Antsfeld, Boris Chidlovskii, and J{\'e}r{\^o}me Revaud.
\newblock Croco v2: Improved cross-view completion pre-training for stereo matching and optical flow.
\newblock In \emph{Proceedings of the IEEE/CVF International Conference on Computer Vision}, pages 17969--17980, 2023.

\bibitem[Wen et~al.(2025)Wen, Trepte, Aribido, Kautz, Gallo, and Birchfield]{FoundationStereo}
Bowen Wen, Matthew Trepte, Joseph Aribido, Jan Kautz, Orazio Gallo, and Stan Birchfield.
\newblock Foundationstereo: Zero-shot stereo matching.
\newblock \emph{arXiv}, 2025.

\bibitem[Wolf et~al.(2024)Wolf, Bracha, and Kimmel]{GS2MESH}
Yaniv Wolf, Amit Bracha, and Ron Kimmel.
\newblock {GS}2{M}esh: Surface reconstruction from {G}aussian splatting via novel stereo views.
\newblock In \emph{European Conference on Computer Vision (ECCV)}, 2024.

\bibitem[Xu et~al.(2023)Xu, Wang, Ding, and Yang]{xu2023iterative}
Gangwei Xu, Xianqi Wang, Xiaohuan Ding, and Xin Yang.
\newblock Iterative geometry encoding volume for stereo matching.
\newblock In \emph{Proceedings of the IEEE/CVF conference on computer vision and pattern recognition}, pages 21919--21928, 2023.

\bibitem[Yang et~al.(2024)Yang, Kang, Huang, Zhao, Xu, Feng, and Zhao]{yang2024depth}
Lihe Yang, Bingyi Kang, Zilong Huang, Zhen Zhao, Xiaogang Xu, Jiashi Feng, and Hengshuang Zhao.
\newblock Depth anything v2.
\newblock \emph{Advances in Neural Information Processing Systems}, 37:\penalty0 21875--21911, 2024.

\bibitem[Yu et~al.(2024)Yu, Sattler, and Geiger]{Yu2024GaussianOF}
Zehao Yu, Torsten Sattler, and Andreas Geiger.
\newblock Gaussian opacity fields: Efficient and compact surface reconstruction in unbounded scenes.
\newblock \emph{ArXiv}, abs/2404.10772, 2024.

\end{thebibliography}
